\documentclass[onefignum,onetabnum]{siamonline190516}
\usepackage{amsmath, amssymb, mathrsfs}
\usepackage{lipsum}
\usepackage{amsfonts}
\usepackage{graphicx}
\usepackage{url}
\usepackage{multirow}
\usepackage{vcell}
\usepackage{xcolor,soul,framed} 

\colorlet{shadecolor}{yellow}

\usepackage{enumitem}
\setlist[enumerate]{leftmargin=.5in}
\setlist[itemize]{leftmargin=.5in}

\usepackage{algorithm}
\usepackage{algorithmic}

\newtheorem{thm}{Theorem}[section]

\newtheorem{rem}[thm]{Remark}

\newtheorem{example}[thm]{Example}

\graphicspath{{./figs/}}

\usepackage{graphicx}
\usepackage{subfigure} 
\usepackage{array}
\usepackage{makecell} 

\makeatletter
\DeclareFontFamily{U}{tipa}{}
\DeclareFontShape{U}{tipa}{m}{n}{<->tipa10}{}
\newcommand{\arc@char}{{\usefont{U}{tipa}{m}{n}\symbol{62}}}%

\newcommand{\arc}[1]{\mathpalette\arc@arc{#1}}

\newcommand{\arc@arc}[2]{%
  \sbox0{$\m@th#1#2$}%
  \vbox{
    \hbox{\resizebox{\wd0}{\height}{\arc@char}}
    \nointerlineskip
    \box0
  }%
}
\makeatother

\usepackage{pgf}

\headers{Image segmentation with topology preserving using ICTM}{Deng, Liu, Wang, and Wang}

\title{Topology preserving Image segmentation using the iterative convolution-thresholding method\thanks{D. Wang is partially supported by National Natural Science Foundation of China (Grant No. 12101524, 12422116), Guangdong Basic and Applied Basic Research Foundation (Grant No. 2023A1515012199), Shenzhen Science and Technology Innovation Program (Grant No. JCYJ20220530143803007, RCYX20221008092843046). X.-P. Wang is partially supported by National Natural Science Foundation of China (Grant No. 12271461, 12426307) and Shenzhen Science and Technology Innovation Program (Grant No. 2024SC0020). D. Wang and X.-P. Wang are partially supported by Guangdong Provincial Key Laboratory of Mathematical Foundations for Artificial Intelligence (2023B1212010001), and Hetao Shenzhen-Hong Kong Science and Technology Innovation Cooperation Zone Project (No.HZQSWS-KCCYB-2024016).}}

 \author{Lingyun Deng\thanks{School of Science and Engineering, The Chinese University of Hong Kong, Shenzhen, Guangdong 518172, P. R. China. (\email{lingyundeng1@link.cuhk.edu.cn}).}
 \and Litong Liu\thanks{Department of Industrial and System Engineering, Georgia Institute of Technology, Atlanta, 30332, United State.
   (\email{lliu607@gatech.edu}).}
   \and Dong Wang\thanks{School of Science and Engineering, The Chinese University of Hong Kong, Shenzhen, Guangdong 518172, P. R. China \& Shenzhen International Center for Industrial and Applied Mathematics, Shenzhen Research Institute of Big Data,  Guangdong 518172, P. R. China
   (\email{wangdong@cuhk.edu.cn}).}
   \and Xiao-Ping Wang\thanks{School of Science and Engineering, The Chinese University of Hong Kong, Shenzhen, Guangdong 518172, P. R. China \& Shenzhen International Center for Industrial and Applied Mathematics, Shenzhen Research Institute of Big Data,  Guangdong 518172, P. R. China
   (\email{wangxiaoping@cuhk.edu.cn}). }} 

\usepackage{amsopn}

\begin{document}

\maketitle

\begin{abstract}
Variational models are widely used in image segmentation, with various models designed to address different types of images by optimizing specific objective functionals. However, traditional segmentation models primarily focus on the visual attributes of the image, often neglecting the topological properties of the target objects. This limitation can lead to segmentation results that  deviate from the ground truth, particularly in images with complex topological structures. In this paper, we introduce a topology-preserving constraint into the iterative convolution-thresholding method (ICTM), resulting in the topology-preserving ICTM (TP-ICTM). Extensive experiments demonstrate that, by explicitly preserving the topological properties of target objects—such as connectivity—the proposed algorithm achieves enhanced accuracy and robustness, particularly in images with intricate structures or noise. 
\end{abstract}



\section{Introduction}

Image segmentation is one of the fundamental tasks in image processing that involves dividing an image into meaningful regions based on properties such as color and shape. It has a wide range of applications, including object recognition, medical imaging, and scene understanding. Segmentation methods can be broadly categorized into traditional approaches and deep learning-based techniques. Among traditional methods, variational approaches have been particularly successful, as they leverage mathematical models to optimize the segmentation process.

Variational methods approach image segmentation by formulating it as an energy minimization problem. The goal is to minimize an energy functional that typically consists of two main components: a fidelity term and a regularization term. The fidelity term ensures that the segmented regions remain consistent with the original image data, while the regularization term imposes smoothness or regularity on the boundaries of the segmented regions. By optimizing this energy functional, variational methods can effectively balance the accuracy of the segmentation with the smoothness of the final contour, leading to more reliable and coherent segmentations. The Mumford-Shah (MS) model\cite{Mumford_1989}, Chan-Vese (CV) model\cite{chan2001active}, and local intensity fitting (LIF) model\cite{li2007implicit} are all examples of commonly used variational methods. Numerical methods can generally be categorized into implicit and explicit methods, depending on whether the boundary is implicitly or explicitly represented. These categories encompass various approaches, including level set methods \cite{chan2001active,osher2001level,osher1988fronts}, phase field methods \cite{bertozzi2006inpainting}, snake models \cite{yezzi1997geometric}, minimal path methods \cite{benmansour2009fast,chen2018minimal}, binary level set methods \cite{lie2006binary,tai2007image}, and graph-based methods \cite{yuan2010study,yuan2010continuous}.

It is well-known that image segmentation algorithms are often sensitive to noise, particularly when objects are located in highly noisy backgrounds, leading to less accurate segmentation results, even with deep learning approaches \cite{milletari2016v,havaei2017brain,minaee2021image,liu2025inverse}. However, in certain cases where images exhibit specific topological properties, these properties can be leveraged to mitigate the influence of noise. In the proposed approach, we integrate a novel topology-preserving geometric deformable model, inspired by the work in \cite{han2003topology}. This method utilizes the concept of simple points from digital topology \cite{Kong_1989, Bertrand_1994} to ensure that the segmentation process preserves the topological structure of the object while reducing the influence of noise. By maintaining the image's topology, the proposed method is capable of eliminating specific types of noise and ensuring that the segmented result remains faithful to the original structure, leading to more accurate segmentation. In particular, when topology preservation is prioritized, imposing the topological constraint can also have effects on inpainting. Even though some models (e.g., snake models) explicitly parametrize the boundaries of segments and thus can automatically preserve topology, they typically require numerous iterations to achieve convergence.

Inspired by the threshold dynamics method for simulating multiphase flow \cite{esedoglu2015threshold}, the iterative convolution-thresholding method (ICTM) \cite{wang2022iterative,wang2017efficient,ma2021characteristic} introduces an efficient approach to image segmentation by using characteristic functions to represent each segment. In this framework, the boundaries of segments are implicitly captured through their respective characteristic functions. ICTM approximates the perimeter term, or regularization term, by utilizing heat kernel convolution, facilitating the minimization of the energy functional. This novel method is efficient, with an optimal computational complexity $O(N\log(N))$ ($N$: number of pixels) at each iteration, robust, and applicable to a variety of segmentation models. It has been explored in combination with deep neural networks \cite{liu2022deep}, integrated with local variance models to improve robustness to noise \cite{liu2023active}, and applied to segmentation tasks with shape constraints \cite{luo2023binary}. Although ICTM has the advantage of efficiency, it fails to keep the topology properties of the object and is relatively sensitive to noises. 

In this paper, we consider the general Potts model \cite{tai2023potts} with the topology preserving constraint and utilize the characteristic function to represent the segment. With the definitions on digital topology, we impose the thresholding step by a prediction-correction step to restrict changes exclusively to points that do not affect the topological properties of the object, which helps to preserve the topology. 

The remainder of this paper is organized as follows. Section~\ref{sec:topology property} characterizes the topological properties of a discrete image, while Section~\ref{sec:model} introduces the topology-preserving variational model for image segmentation. Sections~\ref{sec:m3} and~\ref{sec:decay} present the mathematical derivation of the proposed topology-preserving ICTM and establish its stability. In Section~\ref{sec:app}, we explore the applications of the method to the CV and LIF models, supported by extensive numerical experiments to demonstrate its performance. Finally, conclusions and further discussions are provided in Section~\ref{sec:con}.

\section{Characterization of the topology property}\label{sec:topology property}
Before the introduction of the image segmentation model and numerical methods, we first introduce the definitions of topological properties. Motivated by  \cite{han2003topology}, we introduce the concept of topology number that distinguishes whether the removal of a point will alter the topology of a region enclosed by a boundary. Suppose the region is described by a set of discrete points, then one can divide the points into two groups based on whether its removal will change the topology of the region. To be specific, if the removal of a point from the region will not change the topology, we call the point as a simple point. The simplicity of a point can be determined by identifying its geodesic neighborhoods and topology number. In general, this quantity can characterize whether the iterative scheme will change the topology of the domain enclosed by a contour between two steps.

Consider $\Omega$ to be the domain of the image on pixels, we denote $N_n(x)$ the $n$-Neighborhood of a point $x\in \Omega$ and denote $N_n^*(x) = N_n(x) \backslash x$ where we consider $n \in \{4, 8\}$ for the 2-dimensional case and $n \in \{6, 18, 26\}$ for the 3-dimensional case. For the 2-dimensional case, $N_4(x)$ consists of the set of pixels that are adjacent to $x$ in the north, south, east, and west directions, along with the pixel $x$ itself.  $N_8(x)$ consists the northeast, northwest, southeast, southwest points in addition to the points in $N_4(x)$. Similarly, for the 3-dimensional case, $N_n(x)$ contains the $6$, $18$, or $26$ neighbors of $x$ and $x$ itself, depending on the value of $n$.

In the follows, we denote $X$ and $\bar X = \Omega \setminus X$ to be the foreground and background of the image respectively. Then, for the 2-dimensional case, consider $(n, M) = (4, 8)$, we list the following definitions on the simplicity of a point.

\begin{definition}  
	Let $X \subset \Omega$ and $x\in \Omega$. The {\it geodesic neighborhood} of $x$ with respect to $X$ is:
	 $N_n^1(x,X) = N_n^*(x) \cap X$ and $N_n^2(x,X) = \cup \{N_n(y)\cap N_M^*(x)\cap X, y \in N_n^1(x,X) \}$.
\end{definition}

\begin{definition}[Topology Number]
	Let $X \subset \Omega$ and $x\in \Omega$. Use $C_n(X)$ to denote the set of all n-connected components of $X$. The {\it topology number} of $x$ with respect to $X$ is: $T_n(x,X)=\# C_n(N_n^2(x,X))$ and $T_M(x,X)=\#C_M(N_M^1(x,X))$, where $\#$ denotes the cardinality of the set.
\end{definition}

\begin{definition}[Simplicity Function]
	For $X \subset \Omega$ and $\bar X = \Omega\setminus X$, a point $x\in \Omega$ is called a simple point if $T_n(x,X)=1$ and $T_M(x,\bar X) = 1$. Specifically, we denote the simplicity function by:
     \begin{eqnarray}
    	\mathcal{S}_X(x)=\left\{
    	\begin{aligned}
    		1 & \ \   {\rm if} \  x \ {\rm is \ simple  \ with \ respect \ to } \ X, \\
    		0 & \ \ {\rm otherwise}.
    	\end{aligned}
    	\right.
    \end{eqnarray}
\end{definition}

\begin{figure}[ht!]
\centering
\subfigure[Nonsimple Point]{
\includegraphics[width=0.25\linewidth]{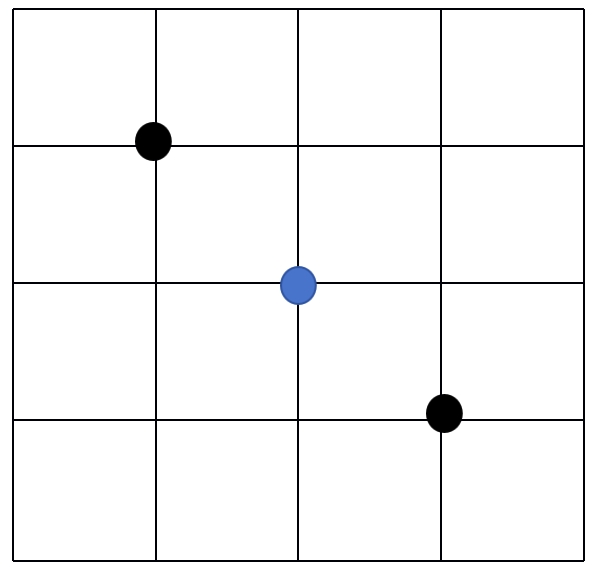}}
\subfigure[Nonsimple Point]{
\includegraphics[width=0.25\linewidth]{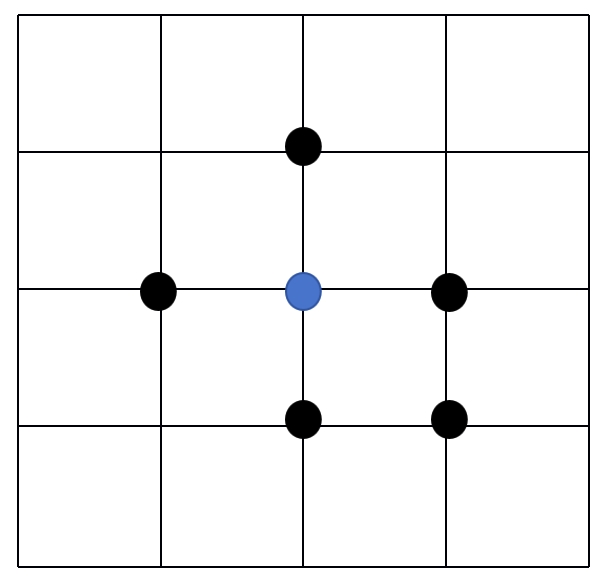}}
\subfigure[Simple Point]{
\includegraphics[width=0.25\linewidth]{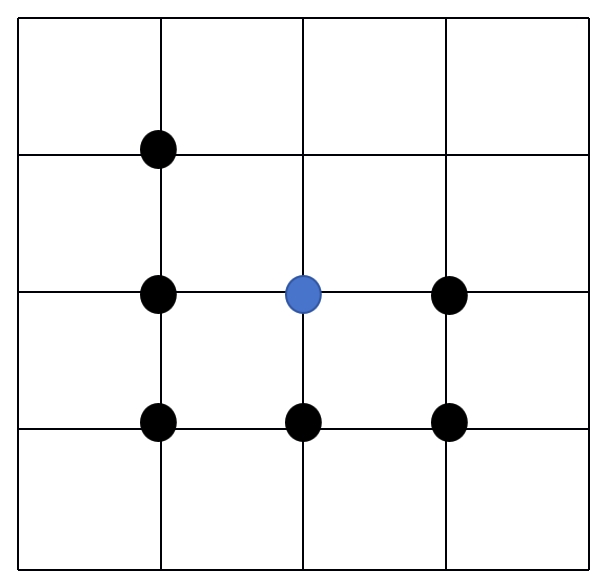}}
\caption{Examples of non-simple and simple points. The removal of the blue point in cases (a) and (b) will change the topology of all black points and the removal of the blue point in case (c) will not affect the topology property. }\label{fig:simple points}
\end{figure}

\begin{example}
In Figure~\ref{fig:simple points}, we use an example to illustrate the definition of simple point and topology number.
Based on above definitions, assume $X$ to be the set of all black points and the blue point $x$, we can calculate the topology number of the blue point in each case:
\begin{itemize}
	\item[1.] Blue point in (a) is not a simple point, it has $T_4(x,X)=2$ and $T_8(x,\bar X) = 1$;
	\item[2.] Blue point in (b) is not a simple point, it has $T_4(x,X)=3$ and $T_8(x,\bar X) = 3$;
	\item[3.] Blue point in (c) is a simple point, with $T_4(x,X)=1$ and $T_8(x,\bar X) = 1$.
\end{itemize}
\end{example}
We can easily observe that, the removal of the blue point in cases (a) and (b) will change the topology of all points and the removal of the blue point in case (c) will not affect the topology property. 

\begin{rem}
In Figure~\ref{fig:simple points}, we consider the case where $(n,M) = (4,8)$. That is, we consider the $4$-connectivity for the foreground and $8$-connectivity for the background.   One can also consider the case where $(n,M) = (8,4)$. In order to avoid a connectivity paradox, different connectivities, $n$ and $M$, must be used in a binary image
comprising the foreground and the background.
\end{rem}

\section{The topology preserving variational model} \label{sec:model}
In this section, we consider a general objective functional in variational models applied to image segmentation:
\begin{align}
\mathcal J = \sum_{i=1}^s \int_{\Omega_i} F_i(f, \Theta_1, \ldots, \Theta_n) \ dx + \lambda \sum_{i=1}^s |\partial \Omega_i|, \label{EnergyGeneral}
\end{align}
where $f: \Omega \rightarrow [0,1]^d$ is the image, $d$ is the number of channels, $\Theta_i = (\Theta_{i,1}, \Theta_{i,2}, \ldots, \Theta_{i,m})$ contains all possible variables or functions used in the fidelity terms, $\lambda$ is a regularization parameter, $\Omega_i, i = 1,2,\cdots, s-1$ denotes the domain of each segment in the foreground, $\Omega_s$ denotes the background, and $|\partial \Omega_i|$ represents the perimeter of the boundary $\partial \Omega_i$. The function $F_i$ is general and can encompass numerous well-known models as special cases, including the Chan-Vese (CV) model \cite{chan2001active,vese2002multiphase}, the locally implicit fitting (LIF) model \cite{li2007implicit,li2008minimization}, the local Gaussian intensity fitting (LGIF) model \cite{wang2009active}, the level set active contour (LSAC) model \cite{zhang2015level}, and even topology optimization models \cite{chen2022efficient}.

We then propose the following constrained minimization problem:
\begin{align} \label{prob:original}
\begin{cases}
& \min\limits_{\Omega_i}   \  \mathcal J \\
& {\rm s.t.} \ n\text{-connected components of} \  \Omega_i \ \text{is fixed for each } i = 1,2,\cdots, s-1 \\
& \ \ \ \ \ \ \text{and}\ M\text{-connected components of} \ \Omega_s \ \text{is fixed}
\end{cases} 
\end{align}
where $(n,M) = (4,8)$ or $(8,4)$.

Problem~\ref{prob:original} in general is an interface related optimization problem with constraints. The approximation of the model is firstly based on the representation of the interface. Corresponding numerical methods are mainly developed along the directions of front tracking methods \cite{bresson2007fast}, phase field based approaches \cite{esedog2006threshold,liu2022two,qiao2022two},  and level set based approaches \cite{chan2001active,vese2002multiphase}.

For explicit methods, such as front-tracking-based approaches, the contour preserves its topology during evolution unless specific techniques are applied to handle topological changes. However, these methods generally lack stability guarantees and, as a result, require relatively small time steps to ensure accuracy. This limitation makes the overall algorithm inefficient, as it necessitates a significantly larger number of iterations. On the other hand, for implicit methods, there is relatively little literature addressing problems with topology constraints. This scarcity is primarily due to the challenges involved in characterizing topology constraints within representations such as phase fields, level sets, and similar frameworks.

Based on the above definition of the topology number, one can derive an explicit index to clearly determine whether the removal of a point will alter the topology of a connected set in a discrete setting. This observation motivates us to use characteristic functions, defined on discrete pixel points, to represent the segments. In doing so, the boundary of the segment is implicitly represented by these characteristic functions and the model is approximated in the framework of the ICTM.

\section{The topology preserving ICTM}\label{sec:m3}
In this section, following the approaches in \cite{wang2017efficient,wang2022iterative}, we propose using characteristic functions of domains to represent the interface. Under this representation, we approximate the minimization problem and develop an efficient iterative method to generate the minimizing sequence preserving the topology of each segment. Without loss of generality, and for the sake of clarity, we consider the two-phase  (i.e.,  $\Omega_1\subset \Omega$: foreground,  $\Omega_2 = \Omega\setminus \Omega_1:$ background) case as an example to illustrate the derivation.

Denote 
\begin{equation}
u(x) := \begin{cases}
1, & x \in \Omega_1, \\
0, & \text{otherwise}.
\end{cases}
\end{equation}
Note that the boundary of $\Omega_1$ is now implicitly represented by $u(x)$.
As  in \cite{esedoglu2015threshold}, when $\tau \ll 1$, the length of $\partial \Omega_1$ can be approximated as:
\begin{equation}
|\partial \Omega_1| \approx \sqrt{\frac{\pi}{\tau}}\int_{\Omega} u G_{\tau}*(1-u) \ dx,
\end{equation}
where $*$ denotes convolution and $G_{\tau}(x) = \frac{1}{4\pi\tau}\exp\left(-\frac{|x|^2}{4\tau}\right)$ is the heat kernel. A rigorous proof of convergence as $\tau\rightarrow 0$ can be referred to Miranda et al. \cite{miranda2007short}. This formula has also been used to model multiphase motions \cite{esedoglu2015threshold}, volume preserving motions \cite{jacobs2018auction}, wetting dynamics \cite{xu2017efficient, wang2019improved}.

The fidelity term in \eqref{EnergyGeneral} can be transformed into integrals over the entire domain $\Omega$ by multiplying the integrand by $u$ and $1-u$ and can be approximated via using the smoothed function of $u$; $G_{\tau_1}(x) *u$. That is,
\begin{align}
\int_{\Omega_1}F_1 \ dx+\int_{\Omega_2}F_2 \ dx & =\int_{\Omega}uF_1+(1-u) F_2 \ dx \nonumber \\
& \approx\int_{\Omega}G_{\tau_1} *u F_1+(1-G_{\tau_1} *u) F_2 \ dx \label{indicatorapprox}\\
& = \int_{\Omega}u G_{\tau_1}*F_1+(1-u) G_{\tau_1} *F_2 \ dx. \nonumber
\end{align}

Therefore, the total objective functional in \eqref{EnergyGeneral} can be approximated as:
\begin{equation}
\mathcal J \approx \mathcal J^\tau(u,\Theta) :=  \mathcal J^\tau_f(u,\Theta) + \mathcal J^\tau_r(u,\Theta),
\end{equation}
where\begin{align}
&\mathcal J^\tau_f(u, \Theta) = \int_{\Omega} u G_{\tau_1}*F_1(f,\Theta) + (1-u)G_{\tau_1}*F_2(f,\Theta) \ dx, \nonumber \\
&\mathcal J^\tau_r(u, \Theta) = \lambda \sqrt{\frac{\pi}{\tau_2}}\int_{\Omega} uG_{\tau_2} * (1-u) \ dx. \nonumber
\end{align}
Then, the solution to problem \eqref{prob:original} can be approximated by finding $u^{\tau,*}$ that minimizes:
\begin{align}
(u^{\tau,*},\Theta^{\tau,*}) = \underset{u\in B,\Theta\in S}{\text{argmin}}\ \mathcal J^\tau(u,\Theta) \label{min:theta}
\end{align}
where
\begin{align}
B := \{u(x)\in BV(\Omega, R) | u(x)\in\{0,1\}, \forall x\in\Omega, \text{$u$ has fixed $n$-connected components}\}.
\end{align}
Here, $BV(\Omega,R)$ represents the space of bounded variation functions, and $S=S_1\times S_2$, where $S_i$ is the admissible sets for $\Theta_i$.

Now, we will use the coordinate descent method to minimize $\mathcal J^\tau(u,\Theta)$, starting from an initial guess $(u_0,\Theta_0)$ and iteratively finding the minimizing sequence in the following order:
\begin{align}
\Theta^0,u^1,\Theta^1,\dots,u^k,\Theta^k,\dots \nonumber
\end{align}
Without loss of generality, assuming that $u^k$ has been calculated, we fix $u^k$ and find the minimizer $\Theta^k$ of $\mathcal J^\tau(u_k,\Theta)$. In other words:
\begin{equation}
\Theta^k = \underset{\Theta\in S}{\text{argmin}}\ \mathcal J^\tau(u^k,\Theta). \label{min: datafitting}
\end{equation}
In image segmentation models, such as CV model \cite{chan2001active,vese2002multiphase}, the LIF model \cite{li2007implicit,li2008minimization}, the LGIF model \cite{wang2009active}, and the level set active contour (LSAC) model \cite{zhang2015level}, $S$ is usually the whole Euclidean space and \eqref{min: datafitting} is readily seen to be a convex optimization problem with global minimizers that exist for every fixed $u^k$. Most of these problems can be solved analytically, while others can be efficiently approximated using Gauss–Seidel iterations.

Denote \begin{align}
\tilde B:= \{u(x)\in BV(\Omega, R) | u(x)\in\{0,1\}, \forall x\in\Omega\}.
\end{align}
It is easy to see that $B \subset \tilde B$. Once $\Theta^k$ is obtained, we solve for $u^{k+1}$ by a prediction and correction step, where the prediction step is considered by finding a candidate to decrease the objective functional in $\tilde B$ and the correction step is to preserve the topology in the subset of the prediction.

The prediction $u^{k+1,*}$ is similar to the thresholding step in the original ICTM solved by
\begin{equation}\label{small problem 1}
u^{k+1,*} = \underset{u\in \tilde B}{\text{argmin}}\ \mathcal J^\tau(u, \Theta^k).
\end{equation}
Note that the set $\tilde B$ contains the extreme points of the convex set $K$:
\begin{equation}
K := \{u(x)\in BV(\Omega, R) | u(x)\in[0,1], \forall x\in\Omega\}\nonumber
\end{equation}
When $\Theta^k$ is fixed, similar to \cite{wang2022iterative}, it can be easily verified that $\mathcal J^\tau(u,\Theta^k)$ is a concave functional. The minimizer of a concave functional on a convex set can always be found at the extreme points of the convex set. By finding the minimum points on the convex set $K$, we can relax the original problem \eqref{small problem 1} and obtain the following equivalent problem:
\begin{equation}
u^{k+1} = \underset{u\in K}{\text{argmin}}\ \mathcal J^\tau(u, \Theta^k).
\end{equation}
Using the fact that a concave functional always locates below its linearization, we solve  the minimization of the linearization of $\mathcal J^\tau(u,\Theta^k)$ at $u^k$,
\begin{equation}\label{eq:linear}
u^{k+1,*} = \underset{u\in K}{\text{argmin}}\ \mathcal L^\tau(u, \Theta^k) :=  \int_\Omega u \varphi^k dx + N^k,
\end{equation}
where $N^k$ is a collection of terms that depend on $u^k$ and $\varphi^k = G_{\tau_1}*F_1(f, \theta^k) - G_{\tau_1}*F_2(f, \theta^k)+ \lambda \sqrt{\frac{\pi}{\tau_2}} G_{\tau_2}*(1-2u^k )$. The problem can then be solved in a point-wise manner,
 \begin{align*}
u^{k+1,*} = \begin{cases} 1, \  \ \text{if}  \  \ \varphi^k < 0, \\
0, \  \ \text{otherwise}.
\end{cases} 
 \end{align*}
Above derivation is similar to the previous ICTM and a lot of numerical experiments in \cite{wang2017efficient,wang2022iterative} indicate that such a choice of $u^{k+1,*}$ usually breaks the topology and is relatively sensitive to noises. In the follows, we consider to find a correction based on $u^{k+1,*}$ such that the topology is preserved as that of $u^k$. 

\subsection{Topology preserving correction}
Comparing the values of $u^{k+1,*}$ and $u^k$ at all pixels, we define two subsets to characterize the pixels are intending to change from background to foreground and from foreground to background as follows,
 \begin{align*}
P_1^k : = \{x\in \Omega| \phi^k(x)<0, u^k(x) = 0\}; \\
P_2^k : = \{x\in \Omega|\phi^k(x)>0, u^k(x) = 1\}.
\end{align*}
We then perform a correction to determine \( \tilde{P}_1^k \subset P_1^k \), where \( u \) transitions from \( 0 \) to \( 1 \), and \( \tilde{P}_2^k \subset P_2^k \), where \( u \) transitions from \( 1 \) to \( 0 \) during the \( (k+1) \)-th iteration. This correction is done while ensuring that the topology of the set defined by \( \chi_{\tilde{P}_1^k} + u^k - \chi_{\tilde{P}_2^k} \) is preserved and remains consistent with that of \( u^k \).

To minimize the objective functional value, we adopt the principle that lager absolute values of $\phi^k$ correspond to a higher probability of accepting a transition. As for points in $P_1^k$, we first sort the values $\phi^k(\tilde{x})$ for all $\tilde{x} \in P_1^k$ in an ascending order. For each point $\tilde{x} \in P_1^k$, we compute the neighborhoods 
\[ N_4^2(\tilde{x}, X), \quad N_8^1(\tilde{x}, \bar{X}),\] 
and the simplicity measure $\mathcal{S}(\tilde{x})$. If $\mathcal{S}(\tilde{x}) = 0$, modifying the value of $u$ at $\tilde{x}$ would violate connectivity. In this case, we set
\[ u^{k+1}(\tilde{x}) = u^k(\tilde{x}).\] Otherwise, we update the value at $\tilde{x}$ by setting:
\[u^{k+1}(\tilde{x}) = u^{k+1,*}(\tilde{x}).\]
As for points in $P_2^k$, we sort the values $\phi^k(\tilde{x})$ for all $\tilde{x} \in P_2^k$ in descending order and repeat the same procedure as described for $P_1^k$, checking neighborhoods and updating $u$ accordingly.

By iterating over all points in $P_1^k$ and $P_2^k$ with the respective sorting orders, the objective functional value is iteratively reduced while maintaining connectivity. The topology preservation checking algorithm is summarized in Algorithm~\ref{alg1} and the whole topology preserving iterative convolution-thresolding method (TP-ICTM) is summarized in Algorithm~\ref{alg2}.

\begin{rem}
In the $k$-th iteration, the correction step involves checking the connected components within the $3 \times 3$ neighborhood of points in $P_1^k$ and $P_2^k$. Typically, the number of pixels in $P_1^k$ and $P_2^k$ is proportional to $N$, where $N^2$ represents the approximate total number of pixels. Consequently, the additional computational complexity of this step is linear in $N$ and can be considered negligible compared to the overall computational complexity of each iteration in ICTM, where the convolution operation requires $O(N^2)$ computational complexity.
\end{rem}

\begin{algorithm}[ht!]
\caption{Topology preservation checking algorithm.}\label{alg1}
\begin{algorithmic}
\REQUIRE $u^k$, $u^{k+1,*}$, and $\phi^k$
\ENSURE  $u^{k+1}$
\STATE Initialize $u^{k+1} = u^{k+1,*}$
\STATE Define $P_1^k : = \{x\in \Omega| \phi^k(x)<0, u^k(x) = 0\}$ and 
$P_2^k : = \{x\in \Omega|\phi^k(x)>0, u^k(x) = 1\}$.
\STATE Sort $P_1^k $ by ascending $\phi^k$ values, and $P_2^k $ by descending $\phi^k$ values
\FOR{$\tilde x \in P_1^k$}
    \STATE Compute $N_4^2(\tilde x, X)$ and $N_8^1(\tilde x, \bar{X})$
    \STATE Determine simplicity $\mathcal{S}(\tilde x)$
    \IF{$\mathcal{S}(\tilde x)=1$}
        \STATE $u^{k+1}(\tilde x) = u^{k+1,*}(\tilde x)$
    \ELSE
        \STATE $u^{k+1}(\tilde x) = u^k(\tilde x)$
    \ENDIF
\ENDFOR
\FOR{$\tilde x \in P_2^k$}
    \STATE Compute $N_4^2(\tilde x, X)$ and $N_8^1(\tilde x, \bar{X})$
    \STATE Determine simplicity $\mathcal{S}(\tilde x)$
    \IF{$\mathcal{S}(\tilde x)=1$}
        \STATE $u^{k+1}(\tilde x) = u^{k+1,*}(\tilde x)$
    \ELSE
        \STATE $u^{k+1}(\tilde x) = u^k(\tilde x)$
    \ENDIF
\ENDFOR
\end{algorithmic}
\end{algorithm}

\begin{algorithm}[!ht]
\caption{Topology preserving ICTM.}
\label{alg2}
\begin{algorithmic}
\REQUIRE Given image $f$, $\tau_1, \tau_2 > 0$; tolerance parameter $tol > 0$;  $u^0=\chi_{\Omega^0}$;.
\ENSURE Final segmentation $u$;
\WHILE{$||u^{k+1}-u^k||_2>tol$}
\STATE 1. $\Theta^k = \underset{\Theta \in \mathcal{S}}{\text{argmin}} \int_{\Omega} G_{\tau_1}*u^k F_1+(1-G_{\tau_1}*u^k) F_2 \ dx.$\;

\STATE 2. Compute $\phi^k(x) = G_{\tau_1} *(F_1^k-F_2^k) + \lambda \sqrt{\frac{\pi}{\tau}} G_{\tau_2} *(1-2u^k).$\;
    
\STATE 3. Thresholding: \begin{equation}\nonumber
        u^{k+1,*}(x) = \begin{cases}1 \ \  \textrm{if}  \ \phi^k(x)\leq 0 ,   \\
0 \ \  \textrm{otherwise}.  \end{cases}
    \end{equation}
\STATE 4. Topology preserving: $u^{k+1} = \textnormal{Algorithm~\ref{alg1}}(u^k, u^{k+1,*}, \phi^k).$\;
    \STATE $k=k+1$.
\ENDWHILE
\end{algorithmic}
\end{algorithm}

%
%
%
%

\section{Monotonicity of the objective functional}
In this section, we demonstrate that, under the proposed thresholding prediction and topology-preserving correction, the objective functional remains monotonically decreasing. This ensures the stability of the algorithm during iterations and guarantees convergence to a stationary point.

\begin{thm}[Stability]\label{thm:stability}
Let $(u^k,\Theta^k)$ be the $k$-th iteration generated
in Algorithm~\ref{alg2}. We have
\[\mathcal J^\tau(u^{k+1},\Theta^{k+1}) \leq \mathcal J^\tau(u^{k},\Theta^{k}),\]
for any $\tau_1$ and $\tau_2$.
\end{thm}

{\begin{proof}

The proof has two parts:
$$\mathcal{J}^\tau(u^{k+1},\Theta^{k})  \leq \mathcal{J}^\tau(u^{k},\Theta^{k}),$$
and
$$\mathcal{J}^\tau(u^{k+1},\Theta^{k+1})  \leq \mathcal{J}^\tau(u^{k+1},\Theta^{k}).$$
The second is a direct consequence of \eqref{min:theta}. Therefore, we only need to prove the first one.

Write $$\mathcal{L}^{\tau}(f,\Theta^k,u^k,u^k) 
= \mathcal{J}^{\tau}(u^k,\Theta^k) 
-  \frac{\lambda\sqrt{\pi}}{\sqrt{\tau}}\int_{\Omega} u^k  G_{\tau_2}*u^k \ dx ,
$$
and 
$$\mathcal{L}^{\tau}(f,\Theta^k,u^k,u^{k+1})  
= \mathcal{E}^{\tau}(u^{k+1},\Theta^k) + \frac{\lambda\sqrt{\pi}}{\sqrt{\tau}} \int_{\Omega} u^{k+1} G_{\tau}*(u^{k+1}-2u^k) \ dx.$$

Because the correction is done by selecting the subsets of $P_1^k$ and $P_2^k$ to change values, we have 
$$\tilde P_1^k \subset P_1^k, \ \ \tilde P_2^k \subset P_2^k$$
where 
\[\tilde P_1^k =\{x|u^{k+1}(x)-u^k(x)=1\},\]
\[\tilde P_2^k =\{x|u^{k+1}(x)-u^k(x)=-1\}.\]

From the definition of $P_1^k$ and $P_2^k$, we have 
\begin{align*}
\mathcal{L}^{\tau}(f,\Theta^k,u^k,u^{k+1}) -\mathcal{L}^{\tau}(f,\Theta^k,u^k,u^k) 
=\int_{\tilde P_1^k} \phi^k(x) \ dx  - \int_{\tilde P_2^k} \phi^k(x) \ dx 
\leq  0
\end{align*}
because $\phi^k(x)< 0$ for all $x \in \tilde P_1^k$ and $\phi^k(x)>0$ for all $x \in \tilde P_2^k$.
\end{proof}}

\begin{rem}
The above theorem is consistent with those of the ICTM \cite{wang2022iterative} and guarantees the convergence of the iteration to a stationary point, even when a correction is applied at each step to preserve the topology.
\end{rem}

\section{Applications}\label{sec:app}
The proposed algorithm can be seamlessly applied to and integrated with various variational image segmentation models, while also maintaining the efficiency and stability characteristics of ICTM. In this section, we only focus on the application into two classical models (i.e., the CV model \cite{chan2001active} and the LIF model\cite{li2008minimization}) to show the performance of the proposed algorithm.

\subsection{Application to the CV model} 
The first application of proposed topology preserving ICTM is to the CV model, 
in which the $F_i$ in Algorithm~\ref{alg2} is $F_i = |c_i-f|^2$ and $i =1,2$ for the 2-segment case. When $u^k$ is fixed, $\int_\Omega G_{\tau_1}*u^k|C_1-f|^2 +  (1-G_{\tau_1}*u^k)|C_2-f|^2\ dx$ is strictly convex with respect to $C_1$ and $C_2$. Direct calculation of stationary points yields:
\begin{equation}
    C_1^k = \frac{\int_\Omega (G_{\tau_1}*u^k) f\ dx}{\int_\Omega G_{\tau_1}*u^k \ dx}, \ \ \ \ \ C_2^k = \frac{\int_\Omega (G_{\tau_1}*(1-u^k)) f\ dx}{\int_\Omega G_{\tau_1}*(1-u^k) \ dx}. \label{CVtheta}
\end{equation}
The corresponding TP-ICTM for CV is summarized in Algorithm~\ref{alg:topology preserving CV}.

\begin{algorithm}[!htbp]
\caption{TP-ICTM for CV}
\label{alg:topology preserving CV}
\begin{algorithmic}
\REQUIRE Given image $f$, $\tau_1, \tau_2 > 0$; tolerance parameter $tol > 0$;  $u^0=\chi_{\Omega^0}$;.
\ENSURE Segmentation of the image $u$;
\WHILE{$||u^{k+1}-u^k||_2>tol$}
    \STATE 1. Compute \begin{equation*}
    C_1^k = \frac{\int_\Omega u^k G_{\tau_1}*f\ dx}{\int_\Omega G_{\tau_1}*u^k \ dx}, \ \ \ \ \ C_2^k = \frac{\int_\Omega (1-u^k) G_{\tau_1}*f\ dx}{\int_\Omega G_{\tau_1}*(1-u^k) \ dx}.\end{equation*}
        
    \STATE 2. Compute $\phi^k(x) = G_{\tau_1} *(F_1^k-F_2^k) + \lambda \sqrt{\frac{\pi}{\tau}} G_{\tau_2} *(1-2u^k).$\;
    
    \STATE 3. Thresholding: \begin{equation}\nonumber
        u^{*}(x) = \begin{cases}1 \ \  \textrm{if}  \ \phi^k(x)\leq 0 ,   \\
0 \ \  \textrm{otherwise}.  \end{cases}
    \end{equation}
    \STATE 4. Topology preserving: $u^{k+1} = \textnormal{KeepConnectivity}(u^k, u^{*}, \phi^k).$ \;
     \STATE $k=k+1$.    
\ENDWHILE
\end{algorithmic}
\end{algorithm}

\begin{rem}
In \eqref{CVtheta}, considering a priodic boundary condition on $\Omega$, direct calculation implies that 
$\int_\Omega G_{\tau_1}*u^k f\ dx =\int_\Omega u^k G_{\tau_1}*f\ dx$, $\int_\Omega G_{\tau_1}*(1-u^k) f\ dx = \int_\Omega (1-u^k) G_{\tau_1}*f\ dx$, $\int_\Omega G_{\tau_1}*u^k \ dx = \int_\Omega u^k \ dx$, and $\int_\Omega G_{\tau_1}*(1-u^k) \ dx = \int_\Omega 1-u^k \ dx$. That means, the approximation in \eqref{indicatorapprox} plays an effective role in smoothing the image by a kernel $G_{\tau_1}$ for the CV model.
\end{rem}

In the follows, we give a comprehensive study of the proposed algorithm using the CV model as an example. We use numerical experiments to show the difference (both the iteration process and final results) between TP-ICTM and ICTM, verification of the monotonical decay of the objective functional, sensitivity to parameters and initial guesses, sensitivity to noises, application to images with patterns.

\subsubsection{Difference in the iterations}\label{sec:diffiteration}

\begin{figure}[h]
\includegraphics[width=0.19\linewidth,clip, trim = 2cm 2cm 2cm 2cm]{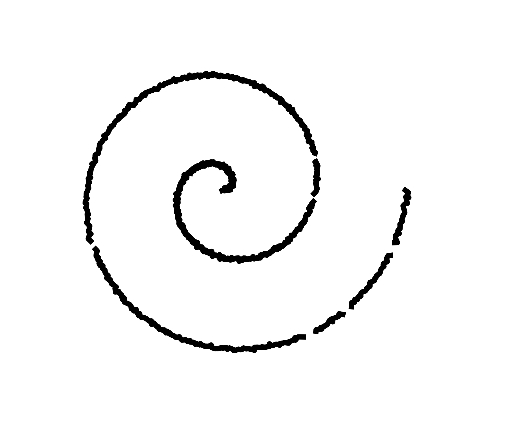}
\includegraphics[width=0.19\linewidth,clip, trim = 2cm 2cm 2cm 2cm]{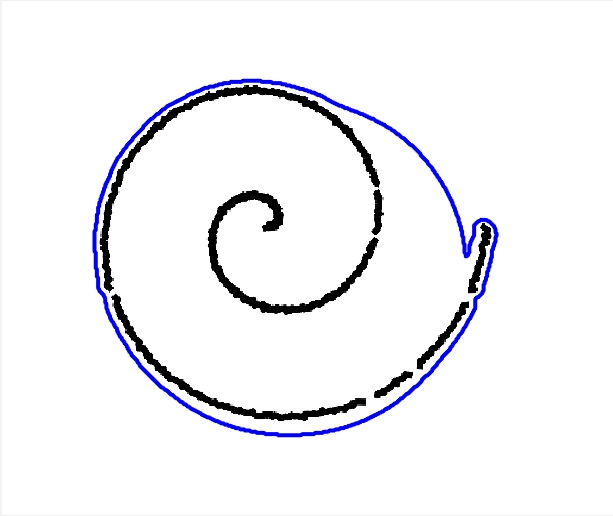}
\includegraphics[width=0.19\linewidth,clip, trim = 2cm 2cm 2cm 2cm]{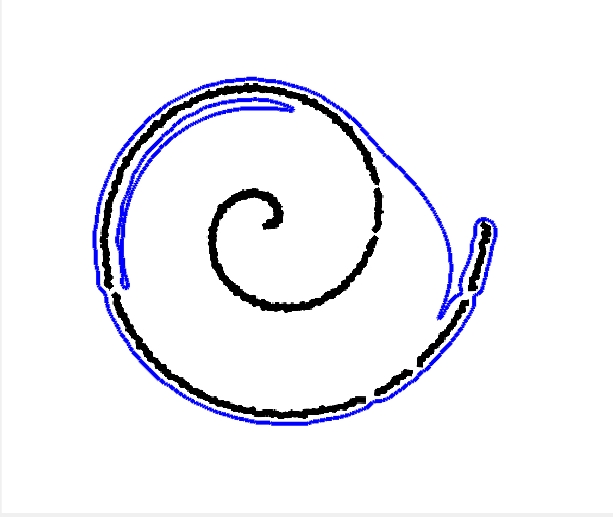}
\includegraphics[width=0.19\linewidth,clip, trim = 2cm 2cm 2cm 2cm]{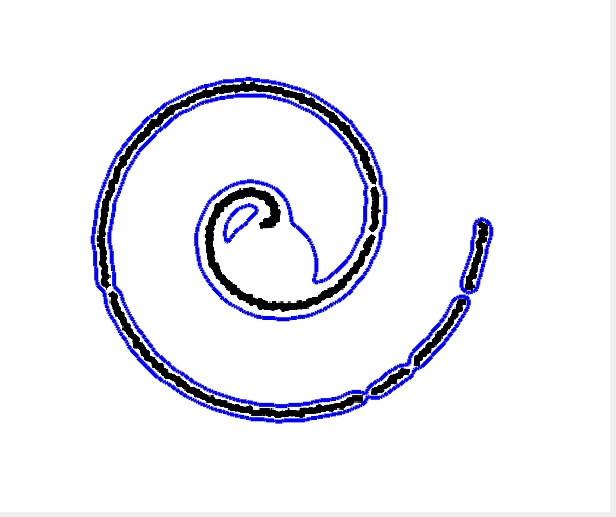}
\includegraphics[width=0.19\linewidth,clip, trim = 2cm 2cm 2cm 2cm]{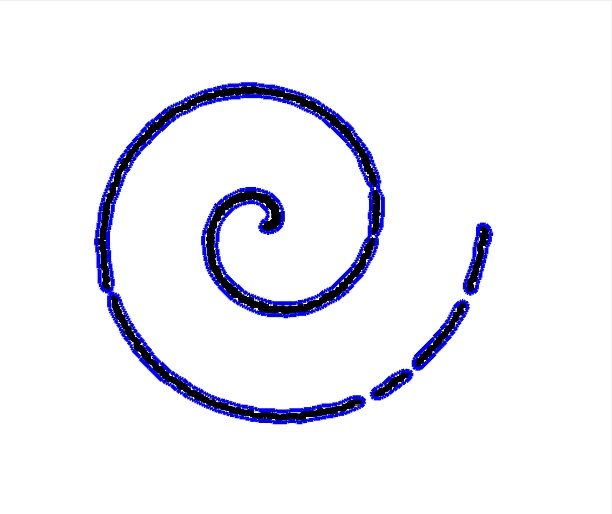}\\
\includegraphics[width=0.19\linewidth,clip, trim = 2cm 2cm 2cm 2cm]{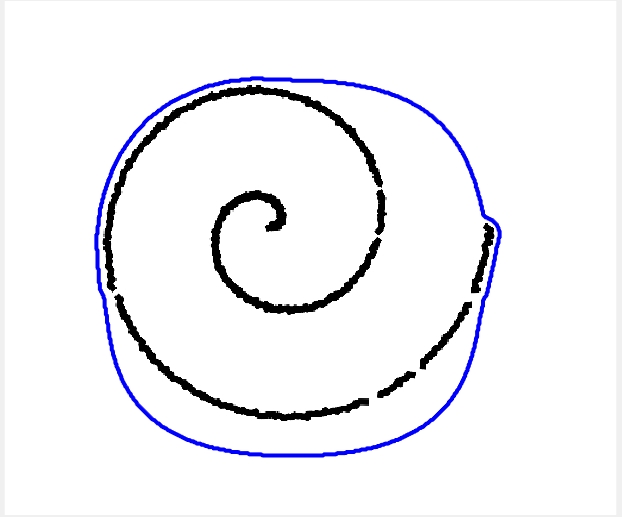}
\includegraphics[width=0.19\linewidth,clip, trim = 2cm 2cm 2cm 2cm]{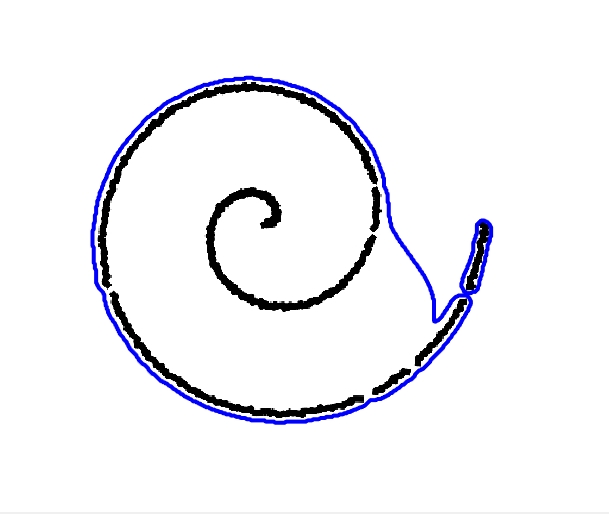}
\includegraphics[width=0.19\linewidth,clip, trim = 2cm 2cm 2cm 2cm]{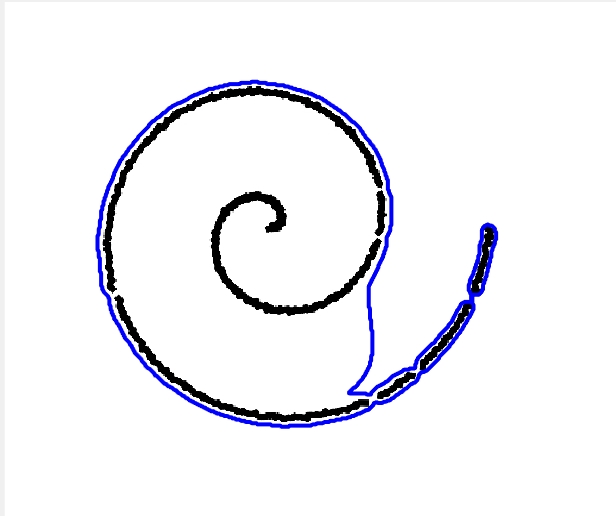}
\includegraphics[width=0.19\linewidth,clip, trim = 2cm 2cm 2cm 2cm]{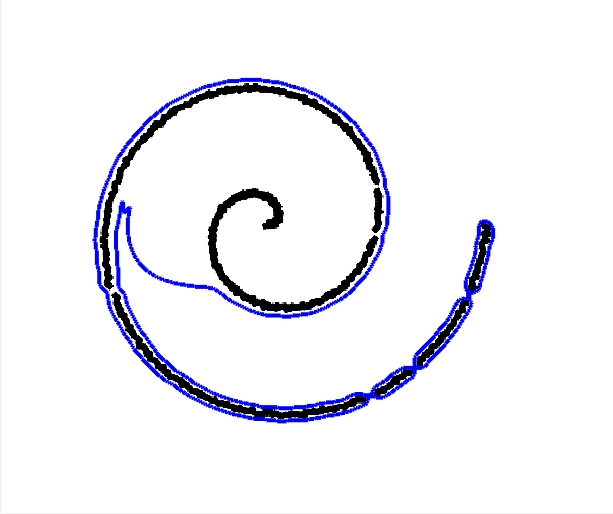}
\includegraphics[width=0.19\linewidth,clip, trim = 2cm 2cm 2cm 2cm]{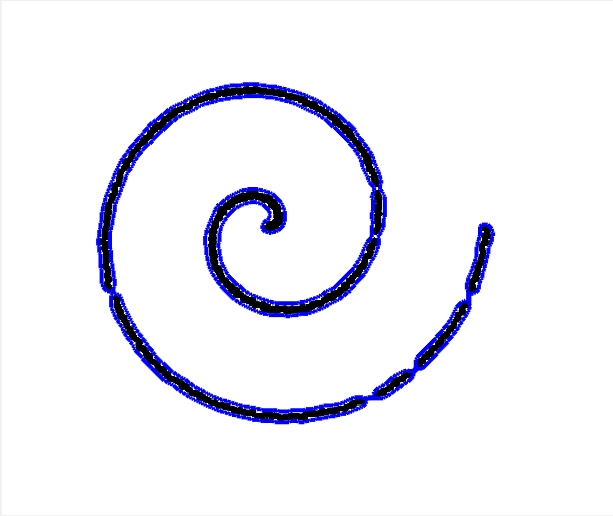}
\caption{First row: selected snapshots of the ICTM for CV. Second row: selected snapshots of the TP-ICTM for CV. See Section~\ref{sec:diffiteration}.} \label{fig:dynamics}
\end{figure}

In this section, we examine the dynamic changes in the iteration when the algorithm is enhanced with corrections for topology preservation. Without these corrections, undesirable topological changes—such as the creation of disconnected regions or the appearance of holes within regions—may arise during the iteration. While such changes can sometimes accelerate the process, they often undermine the desired topological properties. By incorporating corrections for topology preservation, the segmentation process consistently keeps the prior topological attributes of the image, leading to more accurate and reliable results. As shown in Figure \eqref{fig:dynamics}, the TP-ICTM effectively ensures that the segmentation remains fully connected, preserving the topology throughout the process.

\subsubsection{Monotonic decrease of the objective functional} \label{sec:decay}
\begin{figure}[h!]
\centering
\includegraphics[width=0.8\linewidth]{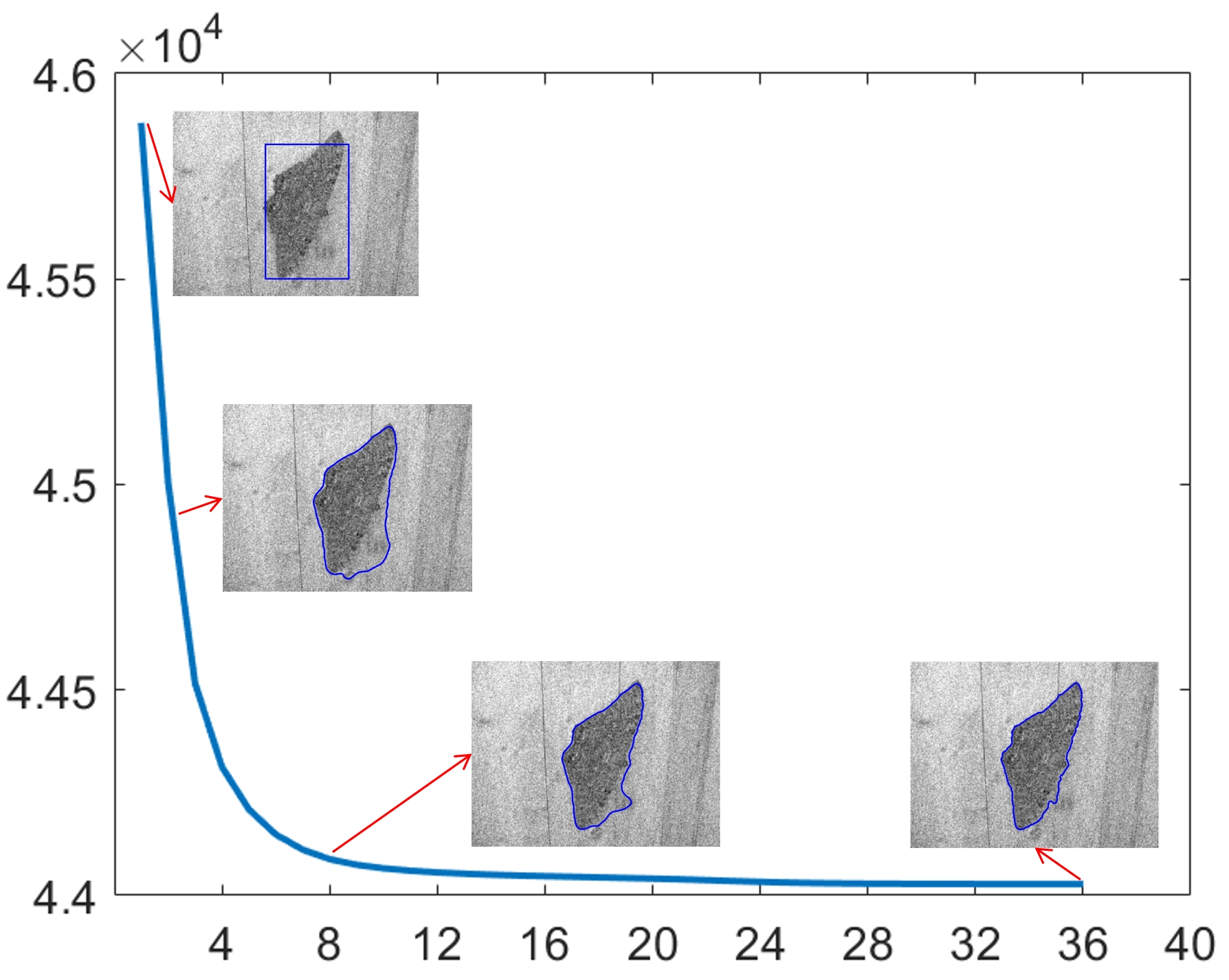}
\caption{Monotonic decrease of the objective functional with some selected snapshots during the iteration process. See Section~\ref{sec:decay}.} \label{fig:objectivefunctionaldecay}
\end{figure}

In this section, we present a numerical example to illustrate the monotonic decay of the objective functional values throughout the iteration process. The plots in Figure~\ref{fig:objectivefunctionaldecay} depict the changes in the objective functional at each iteration step, demonstrating that it consistently decreases over the entire iteration process, which further validates the stability and reliability of the proposed TP-ICTM.
\subsubsection{Parameter tuning}\label{sec:paratuning}
\begin{figure}[h!]
\centering
\includegraphics[width=0.5\linewidth]{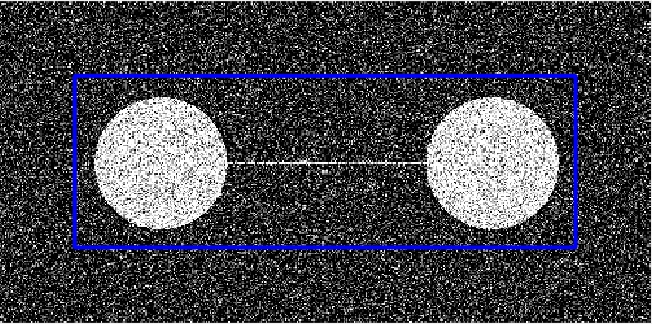}
\caption{Initial guess of a two discs image for parameter tuning in Section~\ref{sec:paratuning}.} \label{fig:twodiscinitial}
\end{figure}

In this section, we demonstrate the impact of adjusting different parameters on the performance of the segmentation algorithm. The parameters in the TP-ICTM for CV are $\tau_1$, $\tau_2$, and $\lambda$. As an example, we consider an image consisting of two discs connected by a thin, narrow line with a same initial guess, as displayed in Figure~\ref{fig:twodiscinitial}.

\begin{figure}[h!]
\centering
\subfigure[$\lambda=0.005$]{
\includegraphics[width=0.3\linewidth]{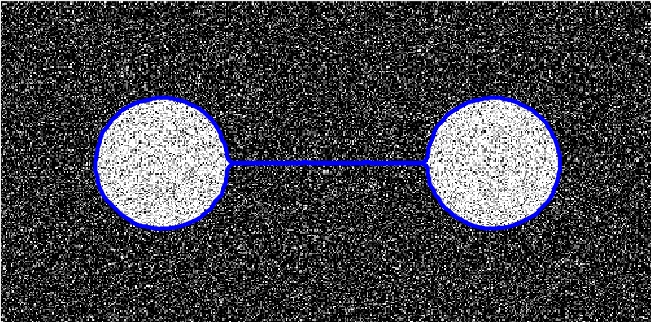}}
\subfigure[$\lambda=0.01$]{
\includegraphics[width=0.3\linewidth]{results/2disk_to.png}}
\subfigure[$\lambda=0.05$]{
\includegraphics[width=0.3\linewidth]{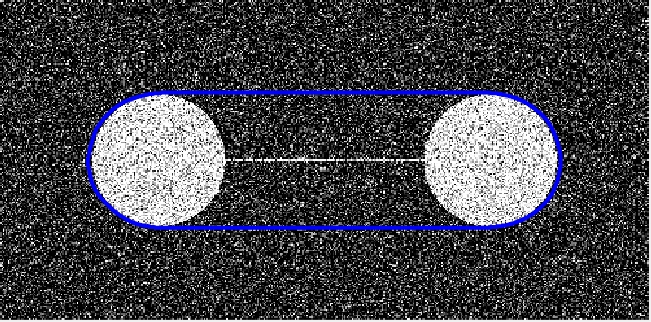}}
\caption{Fix \(\tau_1 = 0.001\) and \(\tau_2 = 0.001\) with $\lambda = 0.005, 0.01,$ and $0.05$. From (a) to (c), the number of iterations required is \(70\), \(128\), and \(418\). See Section~\ref{sec:paratuning}.}
\label{parameter1}
\end{figure}

In Figure \ref{parameter1}, we fix \(\tau_1 = 0.001\) and \(\tau_2 = 0.001\) while incrementally increasing \(\lambda\) from \(0.005\) to \(0.05\). For smaller \(\lambda\) values (\(0.005\) and \(0.01\)), the final results remain similar. When \(\lambda = 0.05\), however, the optimization process becomes dominated by total perimeter minimization, causing the solution to stagnate at a convex shape. In terms of efficiency, the number of iterations required is \(70\), \(128\), and \(418\) for \(\lambda = 0.005\), \(0.01\), and \(0.05\), respectively. This is consistent with the fact that larger values of \(\lambda\) place greater emphasis on the perimeter term, reducing the influence of the image.

\begin{figure}[h!]
\centering
\subfigure[$\tau_2=0.005$]{
\includegraphics[width=0.3\linewidth]{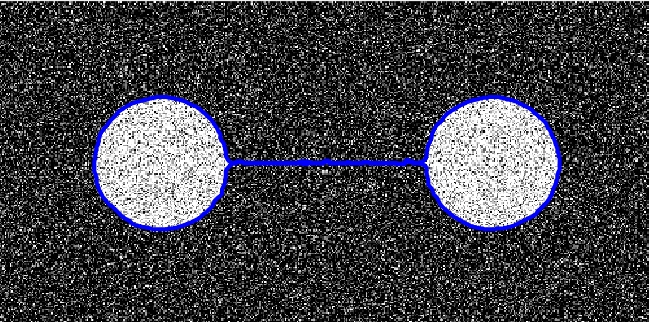}}
\subfigure[$\tau_2=0.01$]{
\includegraphics[width=0.3\linewidth]{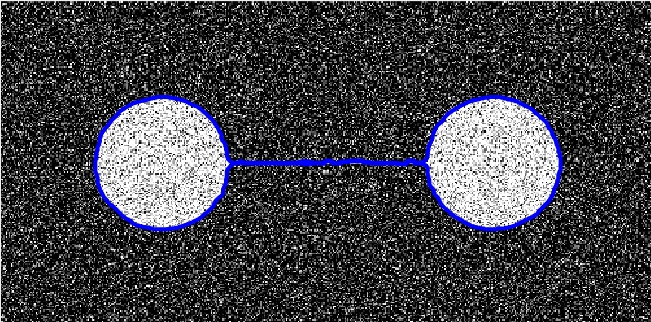}}
\subfigure[$\tau_2=0.1$]{
\includegraphics[width=0.3\linewidth]{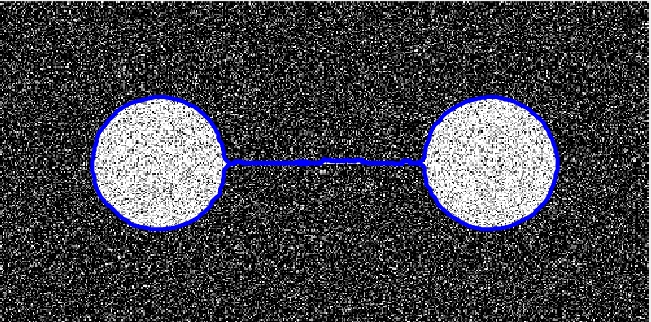}}
\subfigure[Objective functional diagram corresponding to different $\tau_2$]{
\includegraphics[width=0.7\linewidth]{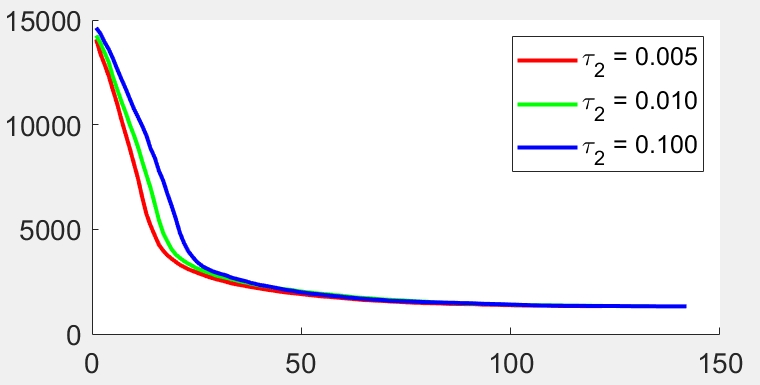}}
\caption{Fix \(\tau_1 = 0.001\) and \(\lambda = 0.001\) with $\tau_2 = 0.005, 0.01,$ and $0.1$. From (a) to (c), the number of iterations required is \(122\), \(133\), and \(142\). (d) is the objective functional value decay curve for different $\tau_2$. See Section~\ref{sec:paratuning}. }
\label{parameter2}
\end{figure}

\begin{figure}[h!]
\centering
\subfigure[$\tau_1=0.001$, snapshot of iteration $13$]{
\includegraphics[width=0.3\linewidth]{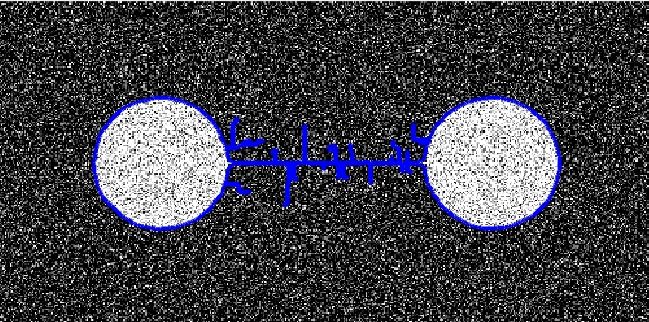}}
\subfigure[$\tau_1=0.003$, snapshot of iteration $13$]{
\includegraphics[width=0.3\linewidth]{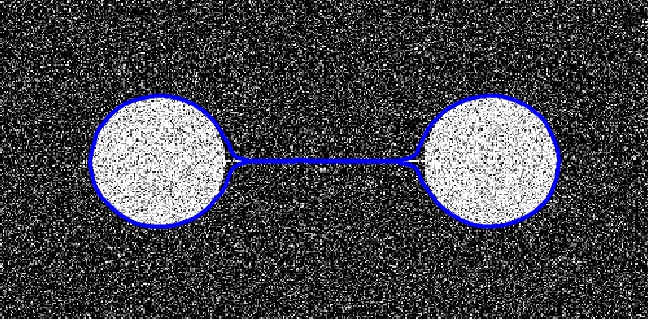}}
\subfigure[$\tau_1=0.005$, snapshot of iteration $13$]{
\includegraphics[width=0.3\linewidth]{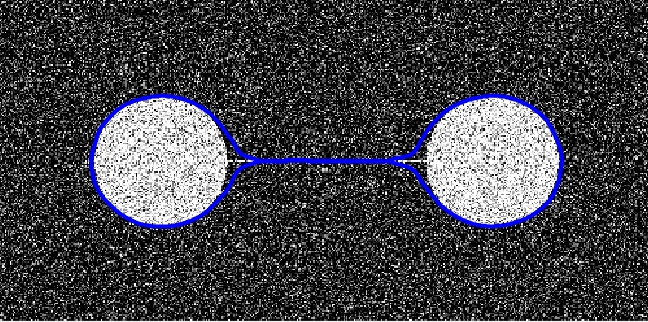}}
\caption{Fix \(\tau_2 = 0.05\) and \(\lambda = 0.05\) with \(\tau_1= 0.001, 0.003,\) and \(0.005\). Snapshots at the \(13\)-th iteration. From (a) to (c), the number of iterations required is \(37\), \(19\), and \(20\).See Section~\ref{sec:paratuning}.}
\label{parameter3}
\end{figure}

In Figure \ref{parameter2}, we fix \(\tau_1 = 0.001\) and \(\lambda = 0.001\) while gradually changing \(\tau_2\) from \(0.005\) to \(0.1\). One can observe that the result is relatively insensitive to the value of $\tau_2$ with a proper choice of $\tau_1$ and $\lambda$. In addition, for $\tau_2 = 0.005, 0.01,$ and $0.1$, the number of iterations required is \(122\), \(133\), and \(142\), respectively. This is also consistent with the fact that \(\tau_2\) is a parameter controlling the approximation of the CV model within the ICTM framework. 
We next investigate the influence of the parameter \(\tau_1\). In this case, we fix \(\tau_2 = 0.05\) and \(\lambda = 0.05\) while varying \(\tau_1\) from \(0.001\) to \(0.005\) (specifically, \(0.001\), \(0.003\), and \(0.005\)). The final results are identical to those shown in Figure~\ref{parameter2}. In Figure~\ref{parameter3}, we present snapshots at the \(13\)-th iteration. It can be observed that larger values of \(\tau_1\) (e.g., cases (b) and (c)) result in a smoother iteration process, avoiding the "cracks" introduced by image noise, as seen in case (a). In terms of efficiency, using a relatively small value of 
$\tau_1$ may require more iterations to achieve convergence, allowing the cracks to evolve internally.
\subsubsection{Sensitivity to initial guesses}\label{sec:sensitivitytoinitialguess}
\begin{figure}[ht!]
\centering
\includegraphics[width=0.2\textwidth,clip, trim = 7cm 2cm 7cm 2cm]{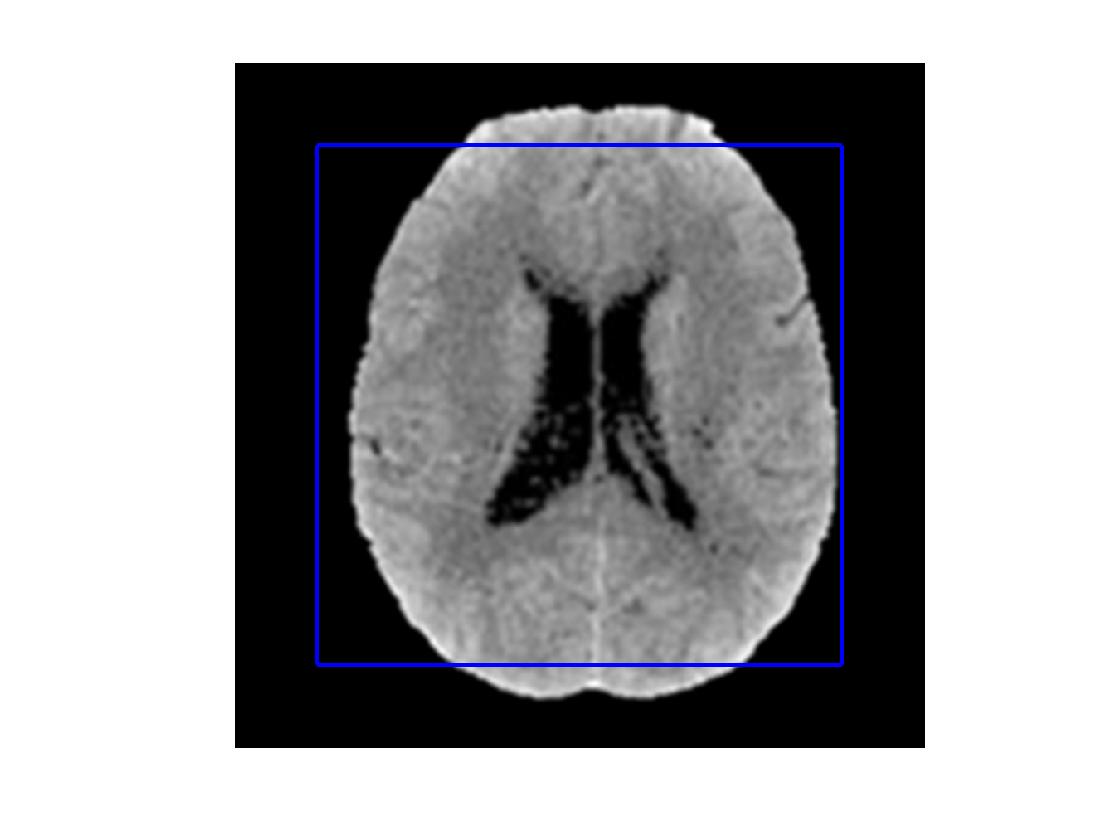}
\includegraphics[width=0.2\textwidth,clip, trim = 7cm 2cm 7cm 2cm]{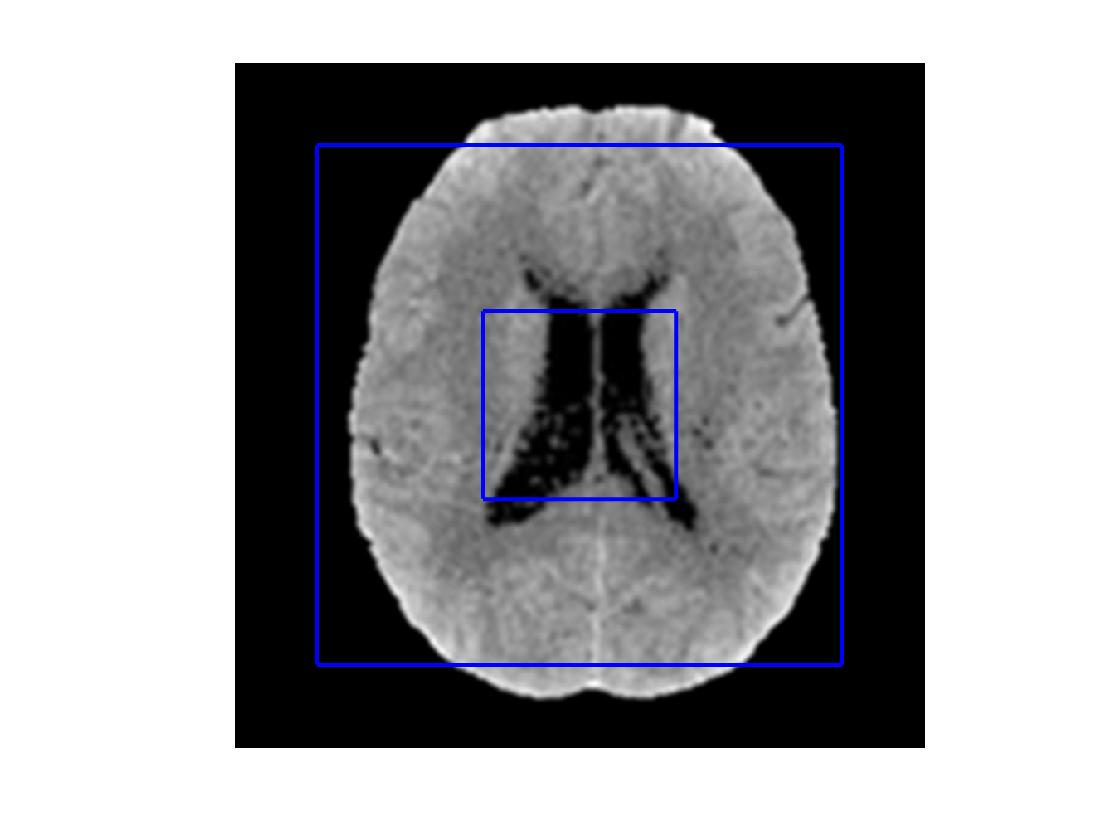}
\includegraphics[width=0.2\textwidth,clip, trim = 7cm 2cm 7cm 2cm]{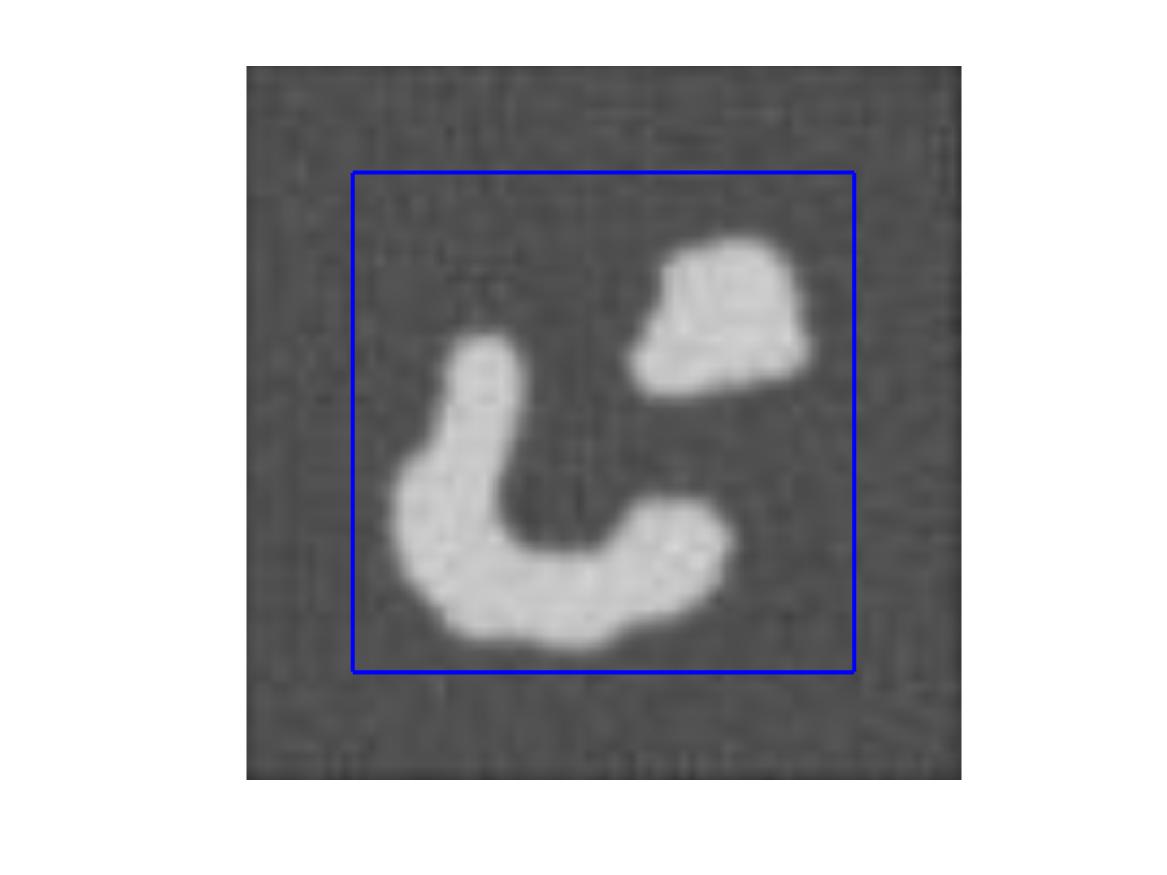}
\includegraphics[width=0.2\textwidth,clip, trim = 7cm 2cm 7cm 2cm]{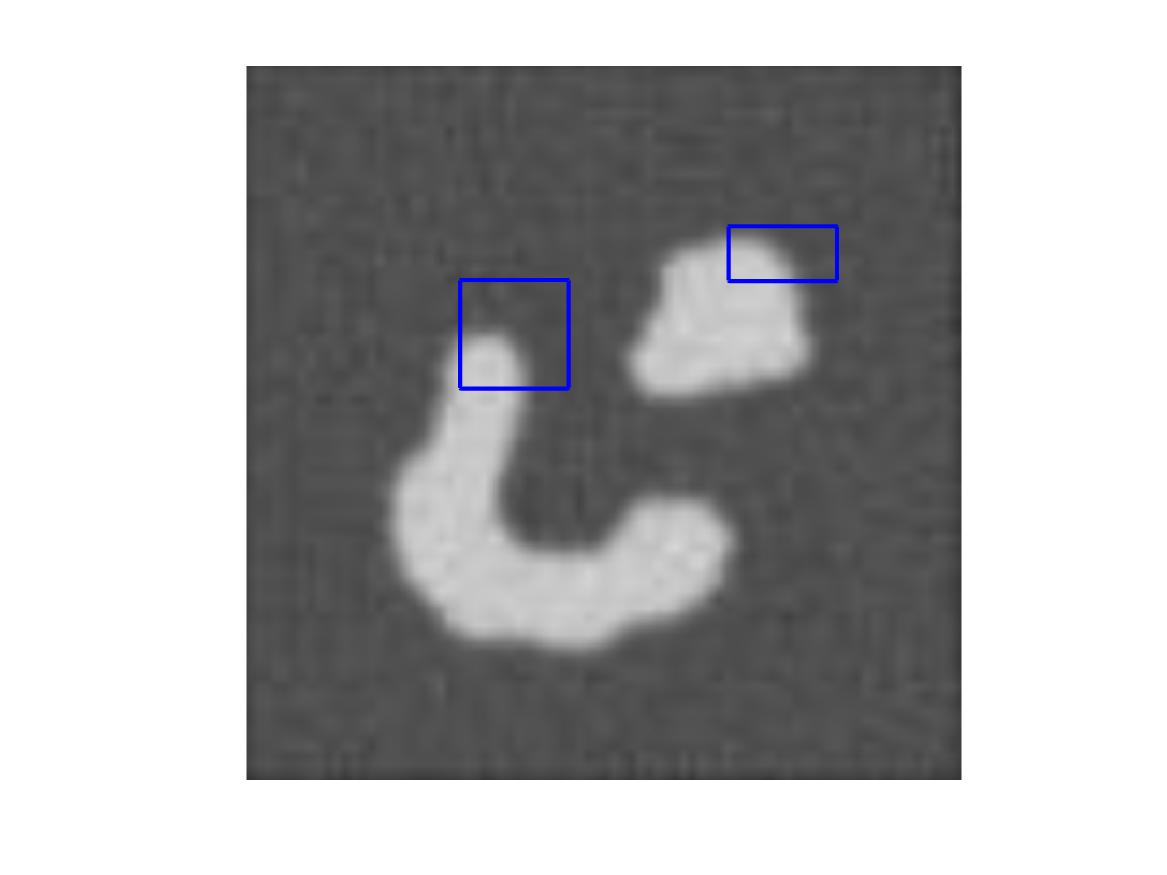}
\includegraphics[width=0.2\textwidth,clip, trim = 7cm 2cm 7cm 2cm]{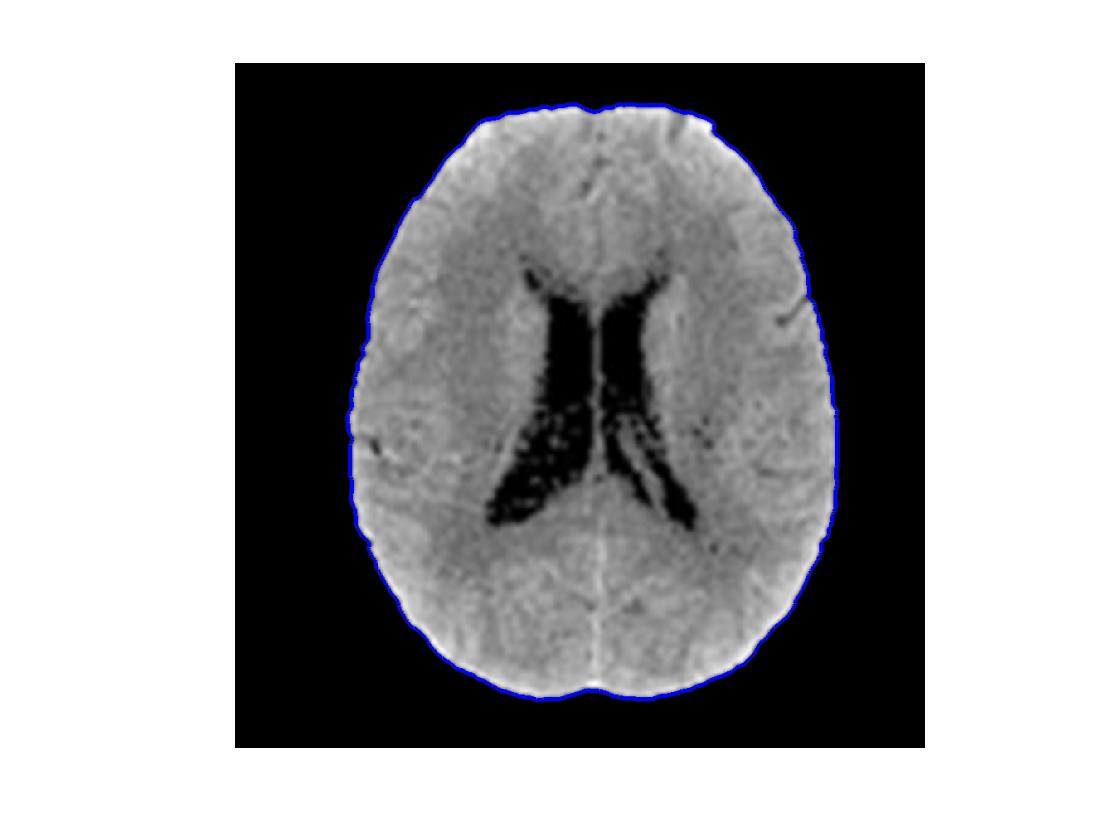}
\includegraphics[width=0.2\textwidth,clip, trim = 7cm 2cm 7cm 2cm]{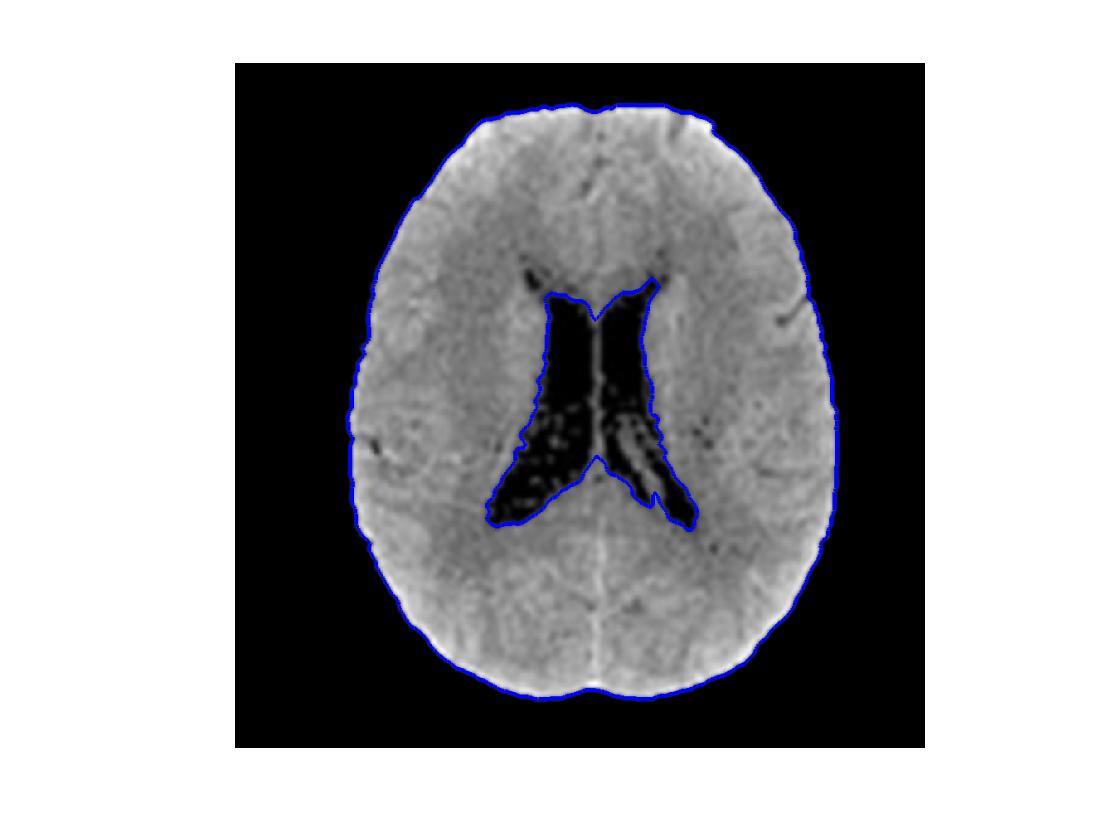}
\includegraphics[width=0.2\textwidth,clip, trim = 7cm 2cm 7cm 2cm]{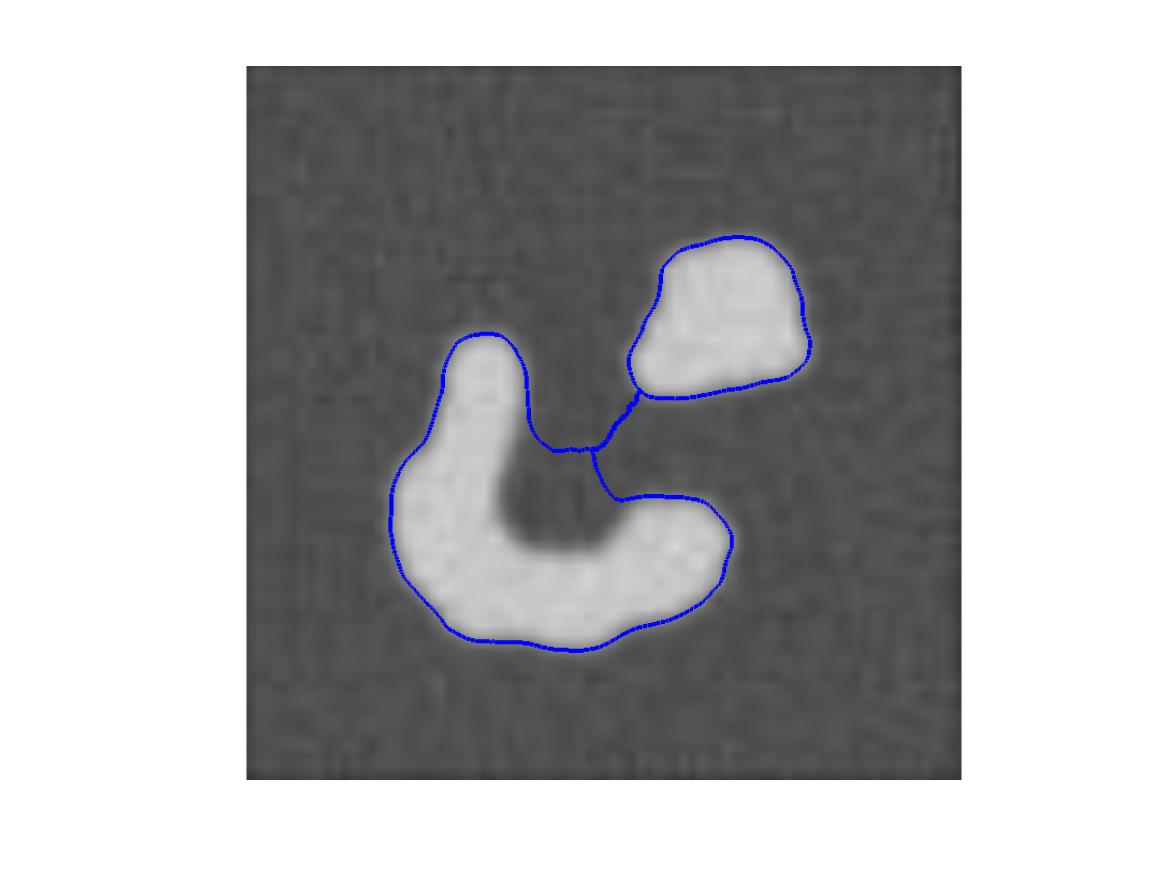}
\includegraphics[width=0.2\textwidth,clip, trim = 7cm 2cm 7cm 2cm]{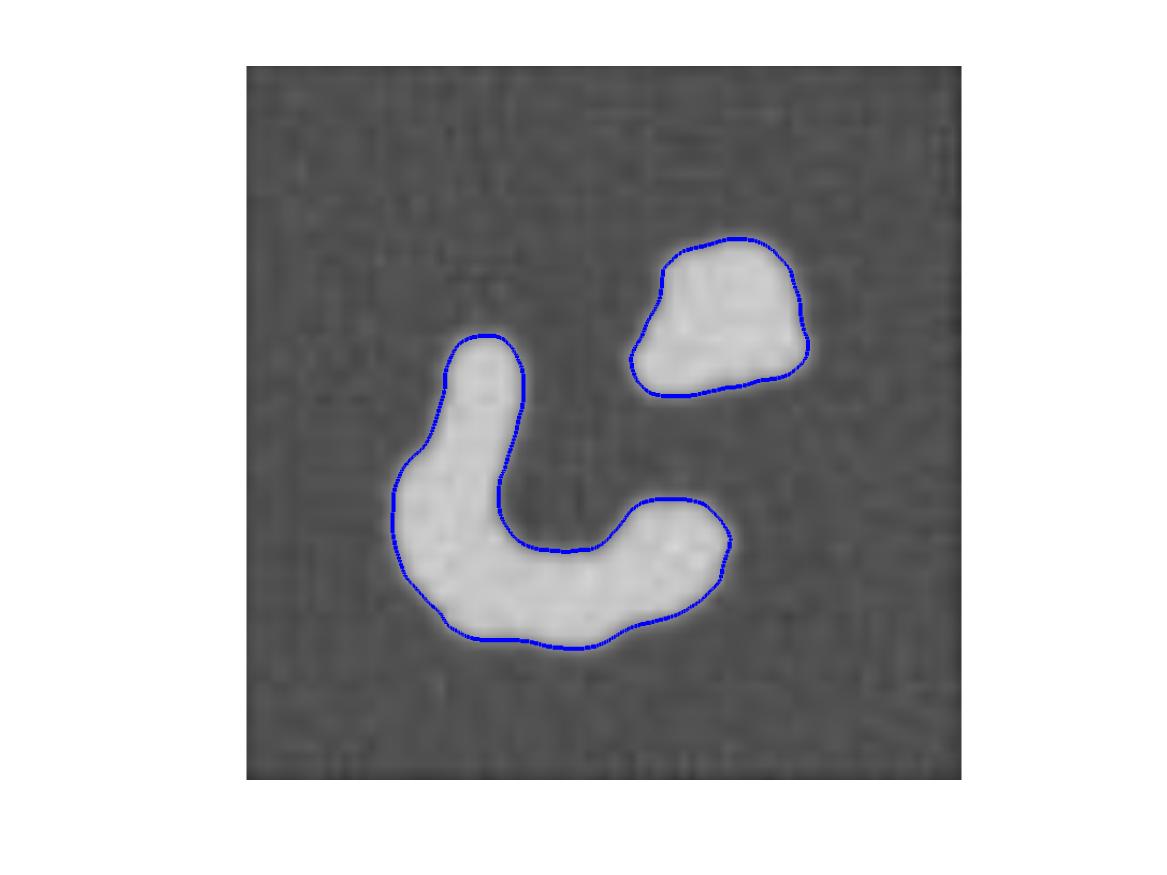}
\caption{Results obtained with different initial guesses. See Section~\ref{sec:sensitivitytoinitialguess}.}\label{fig:images of two subjects}
\end{figure}

\begin{figure}[ht!]
\centering
\includegraphics[width=0.2\textwidth,clip, trim = 7cm 2cm 7cm 2cm]{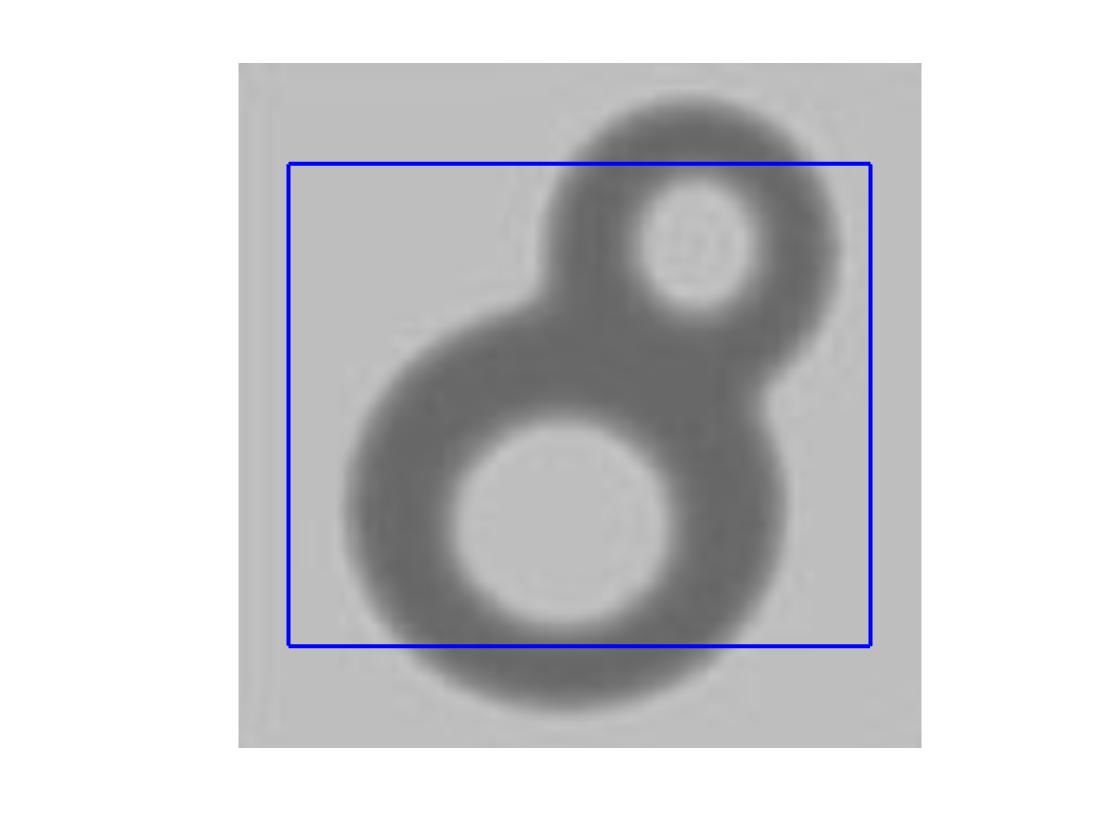}\ \
\includegraphics[width=0.2\textwidth,clip, trim = 7cm 2cm 7cm 2cm]{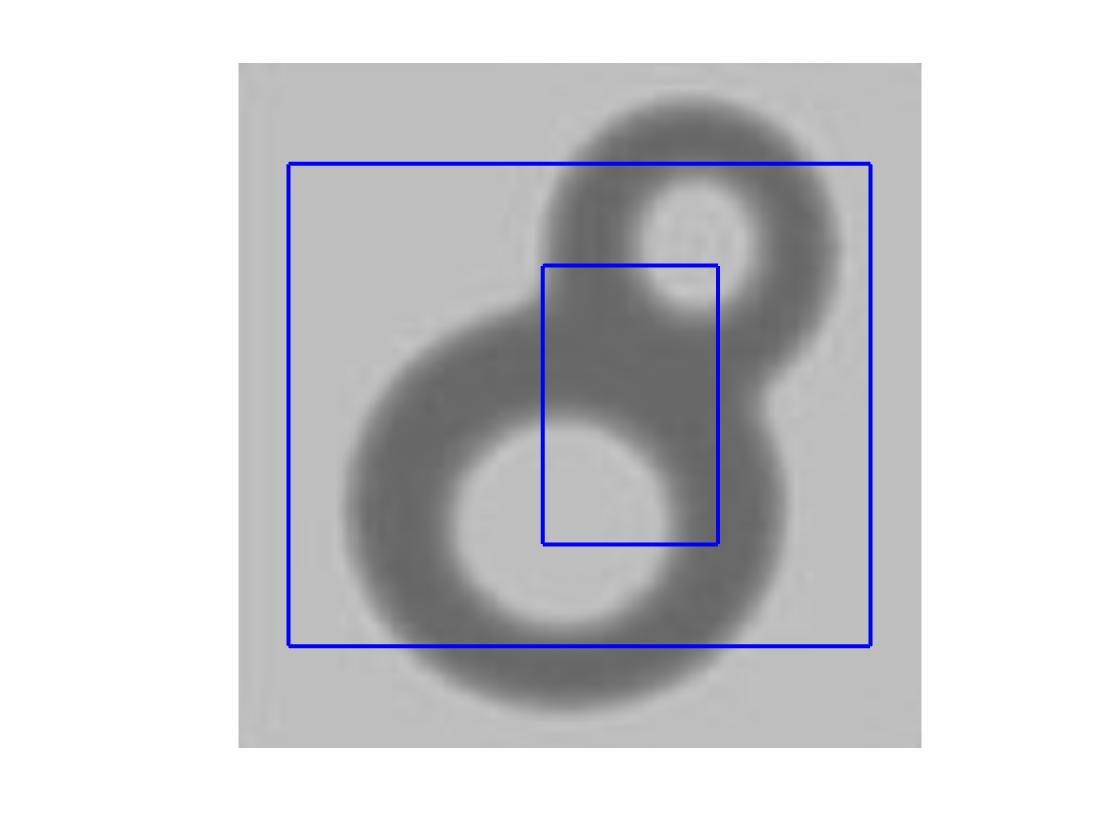}\ \
\includegraphics[width=0.2\textwidth,clip, trim = 7cm 2cm 7cm 2cm]{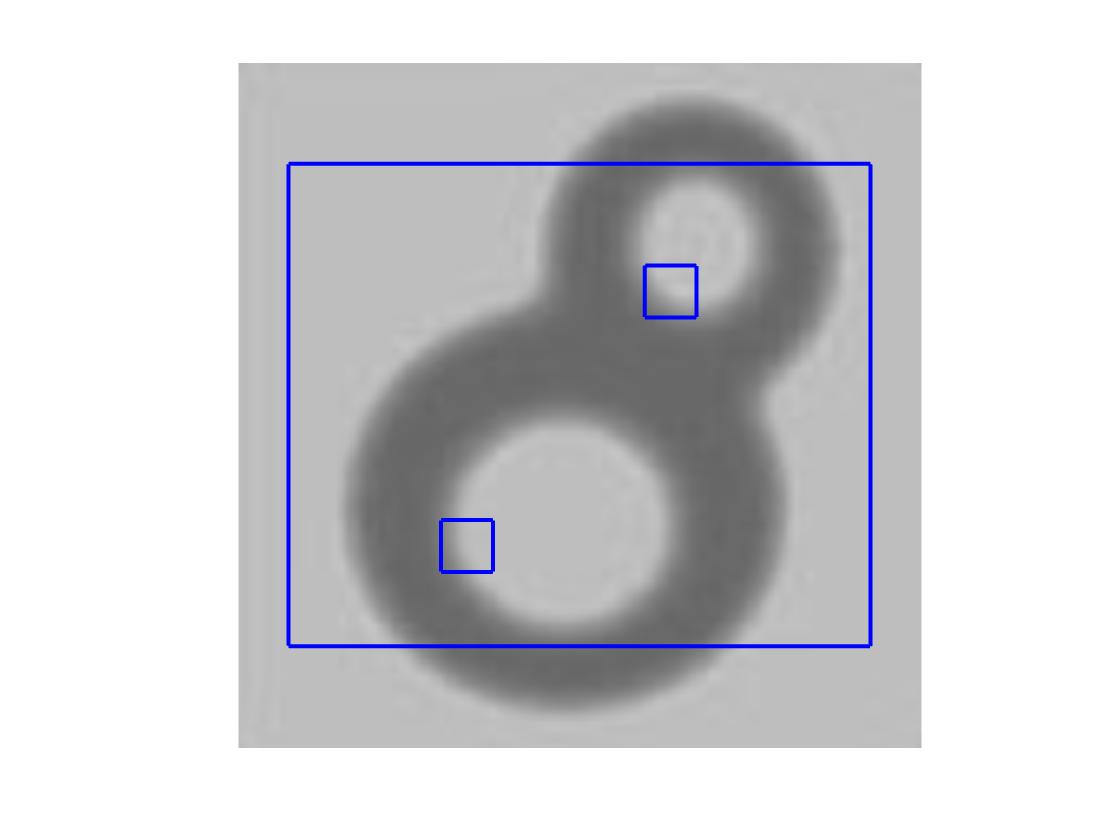}\\
\includegraphics[width=0.2\textwidth,clip, trim = 7cm 2cm 7cm 2cm]{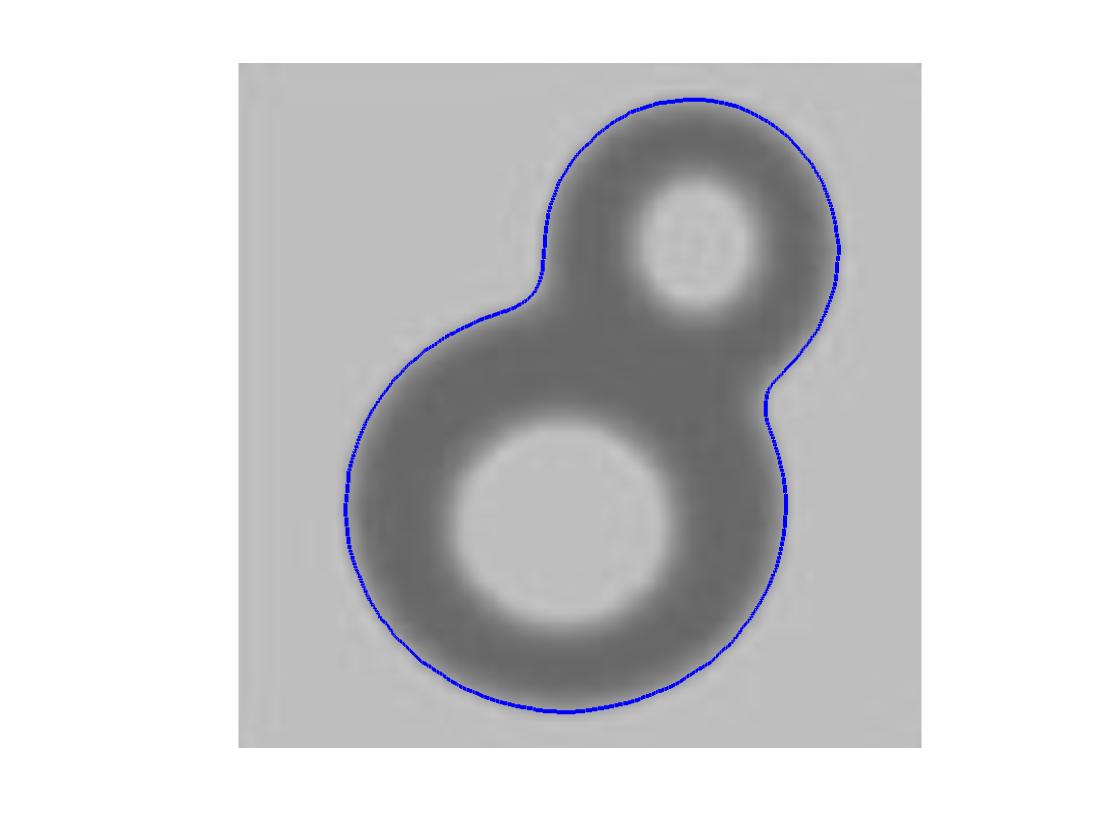}\ \
\includegraphics[width=0.2\textwidth,clip, trim = 7cm 2cm 7cm 2cm]{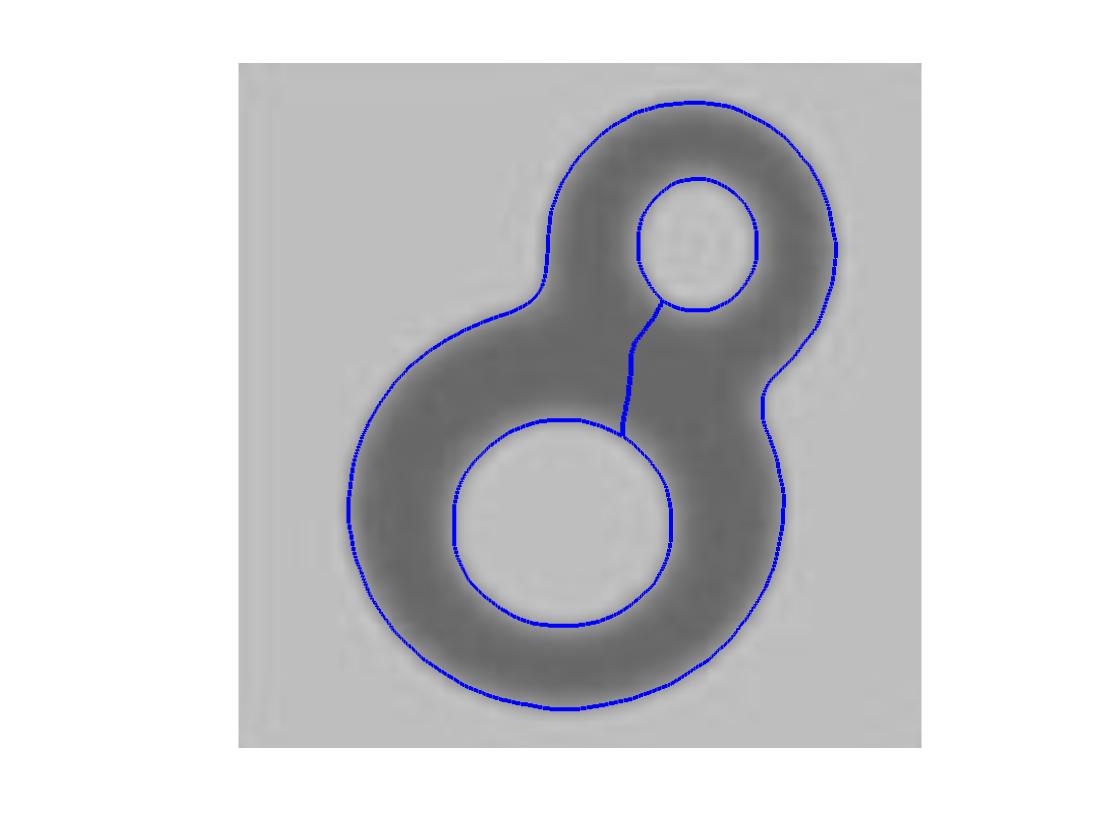}\ \
\includegraphics[width=0.2\textwidth,clip, trim = 7cm 2cm 7cm 2cm]{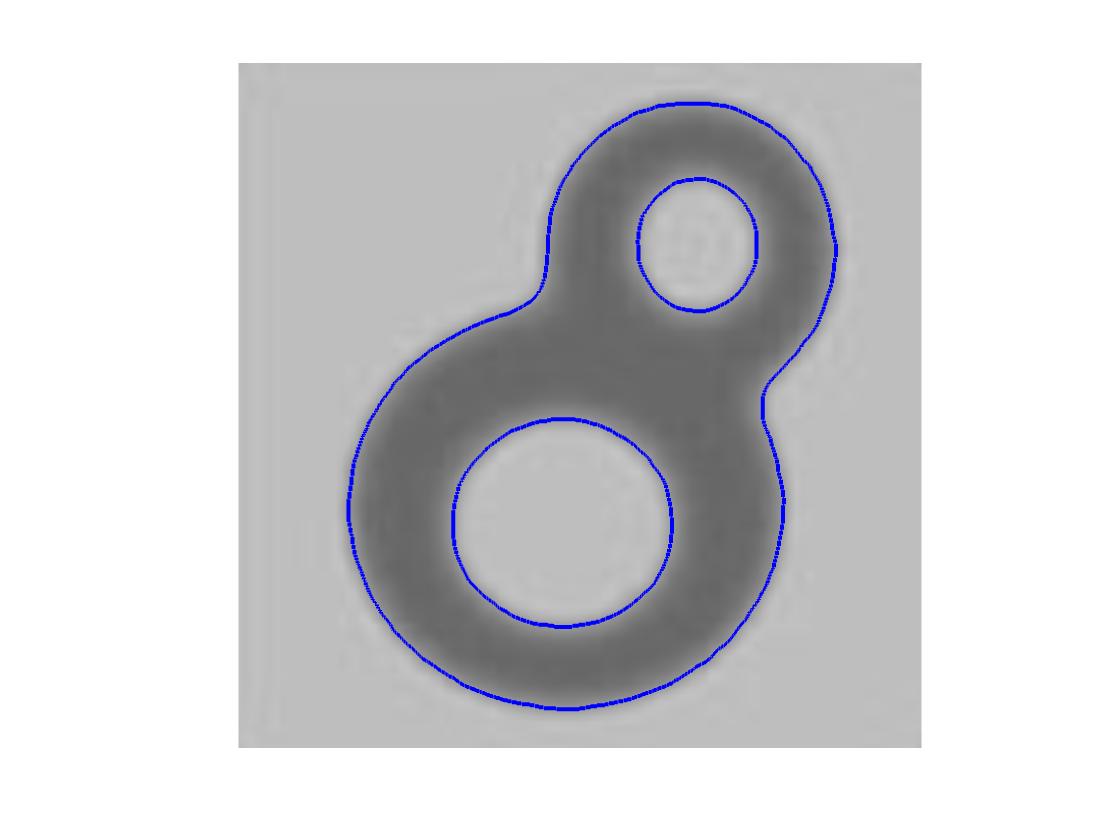}
\caption{Results obtained with different initial guesses. See Section~\ref{sec:sensitivitytoinitialguess}.}\label{fig:images of three subjects}
\end{figure}

In Figure~\ref{fig:images of two subjects} and \ref{fig:images of three subjects}, we use different initial guesses to demonstrate the topology preservation property of the proposed algorithm. Parameters are $\tau_1 = 0$, $\tau_2=0.005$ and $\lambda = 0.01$ in Figure~\ref{fig:images of two subjects} and $\tau_1 = 0$, $\tau_2=0.01$ and $\lambda = 0.005$ in Figure~\ref{fig:images of three subjects}.

In both figures, it is evident that the results depend on the topology number determined by the initial contours. For instance, in Figure~\ref{fig:images of two subjects}, if a square is used as the initial guess, the result will be a connected region without holes. However, if specific topological constraints are imposed (e.g., requiring two disconnected regions or a region with a hole), the results will align with these constraints effectively. Similarly, in Figure~\ref{fig:images of three subjects}, initial guesses with no hole, one hole, and two holes yield results that align with the respective topological constraints. This, in some sense, demonstrates that the proposed TP-ICTM is particularly effective when prior knowledge of topological constraints is available.

\subsubsection{Comparisons to ICTM \cite{wang2022iterative}} \label{sec:comp2ICTM}
\begin{figure}[ht!]
\centering
\begin{tabular}{|c|c|c|c|}
\hline
Initial guess & ICTM-CV & ICTM-LIF & TP-ICTM-CV\\
\hline
\includegraphics[width=0.18\linewidth]{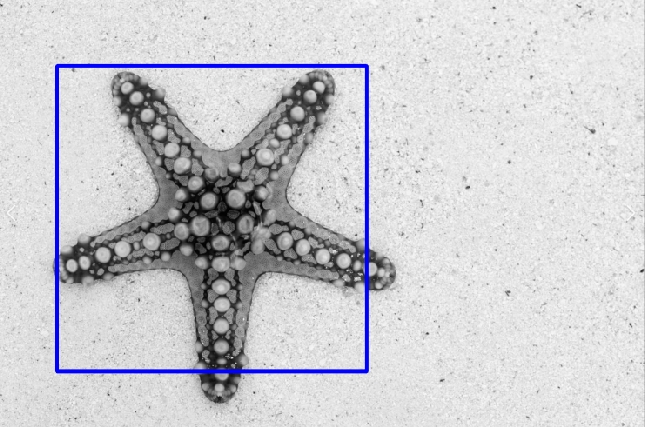}&  
\includegraphics[width=0.18\linewidth]{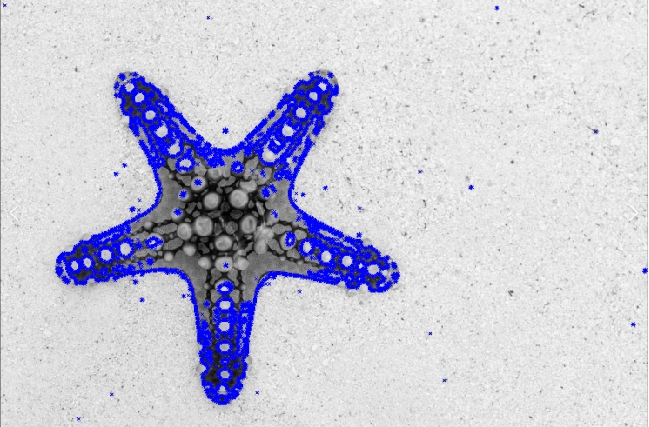}&
\includegraphics[width=0.18\linewidth]{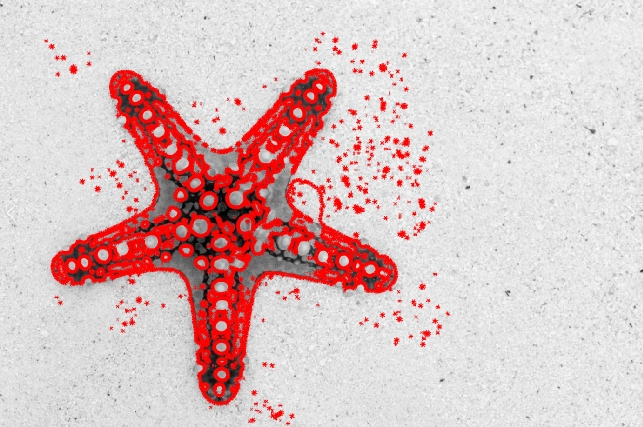}&
\includegraphics[width=0.18\linewidth]{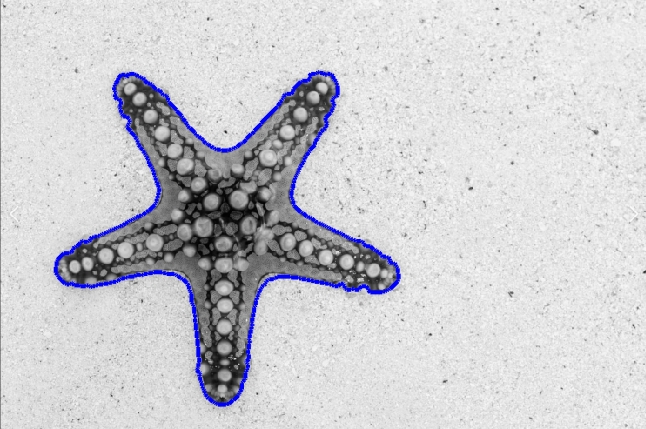}\\
\hline  
\includegraphics[width=0.18\linewidth]{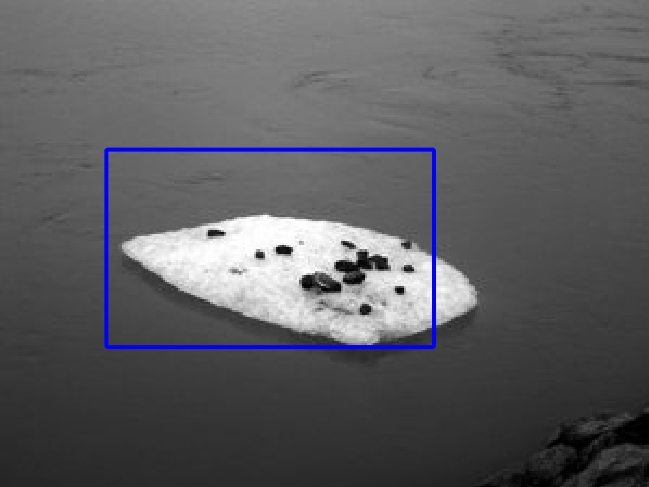}&
\includegraphics[width=0.18\linewidth]{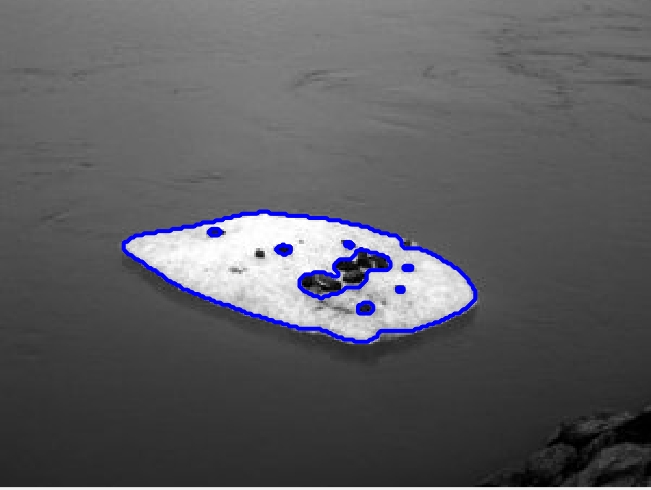}&
\includegraphics[width=0.18\linewidth]{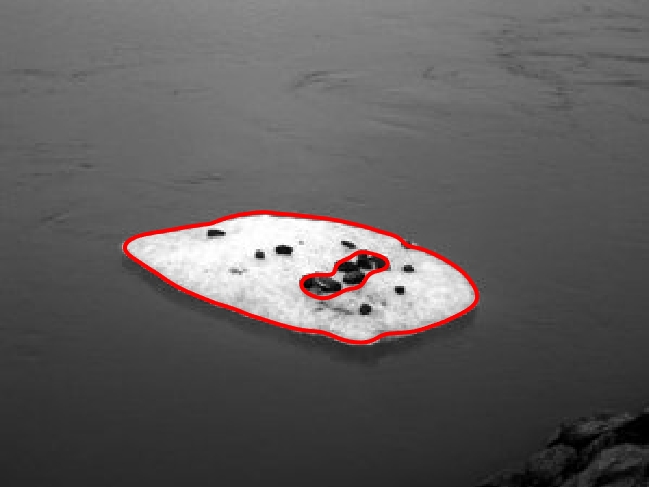}&
\includegraphics[width=0.18\linewidth]{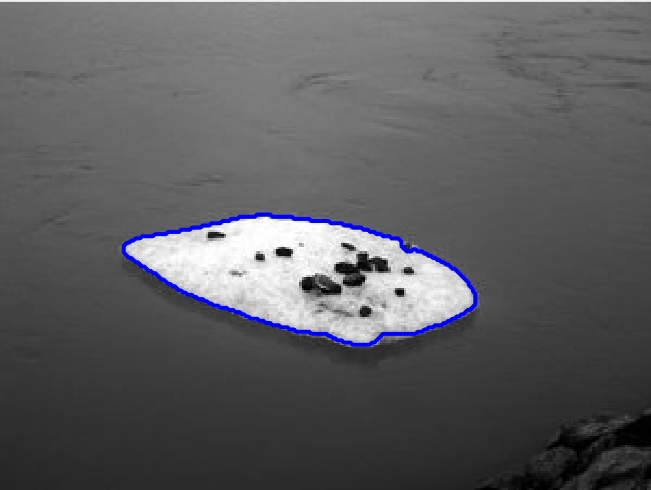}\\
\hline
\includegraphics[width=0.18\linewidth]{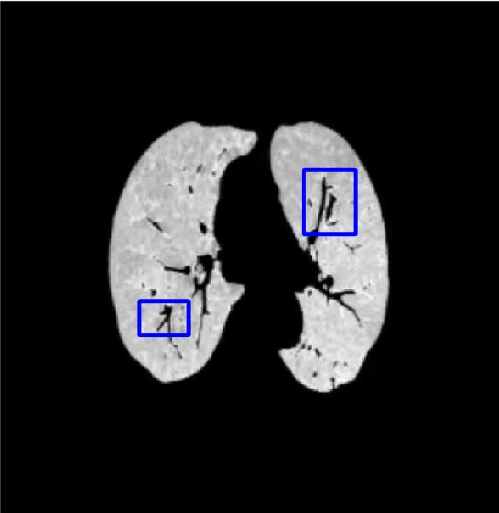}&  
\includegraphics[width=0.18\linewidth]{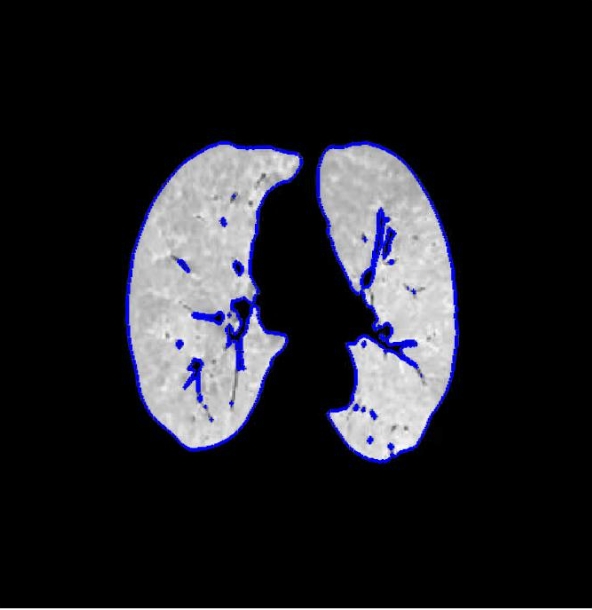}&
\includegraphics[width=0.18\linewidth]{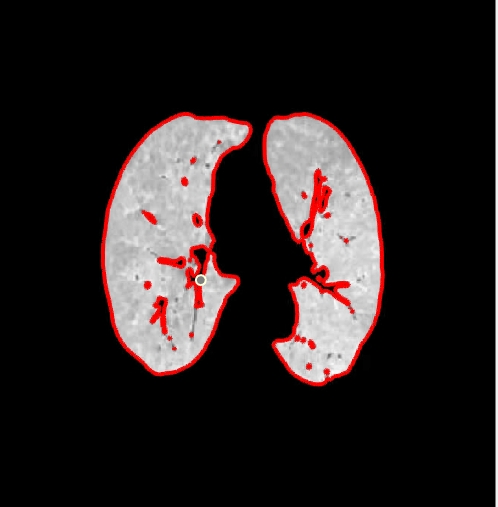}&
\includegraphics[width=0.18\linewidth]{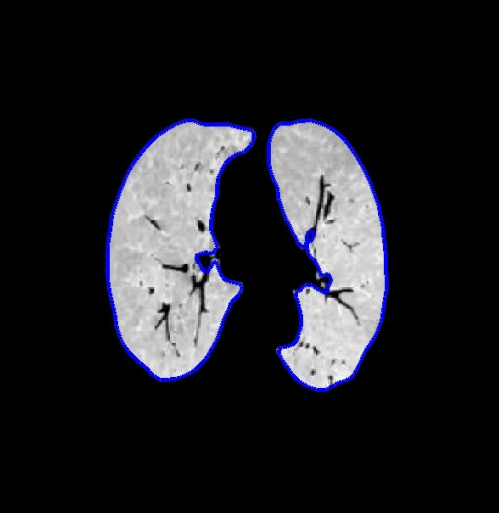}\\
\hline
\includegraphics[width=0.18\linewidth]{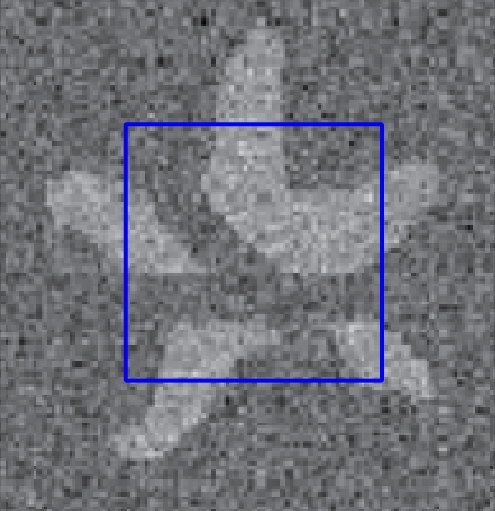}&
\includegraphics[width=0.18\linewidth]{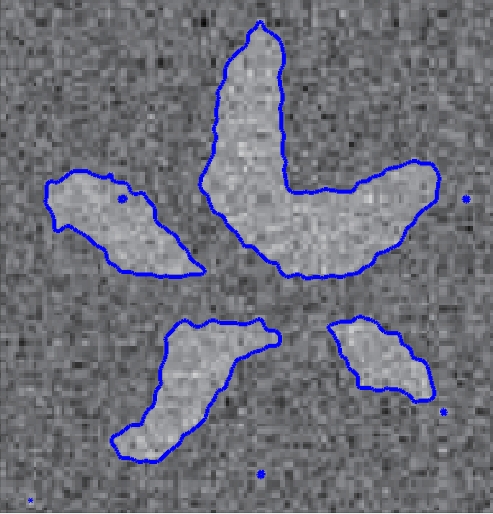}&
\includegraphics[width=0.18\linewidth]{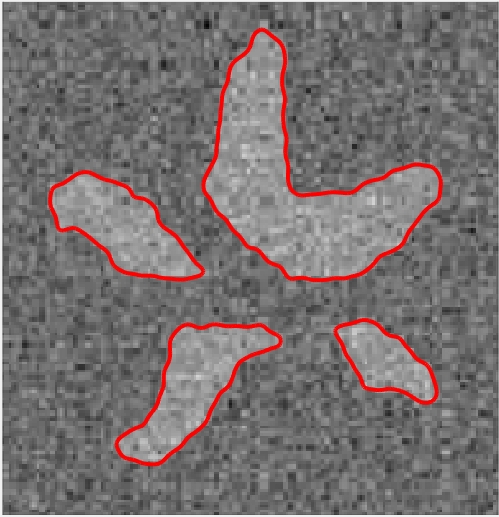}&
\includegraphics[width=0.18\linewidth]{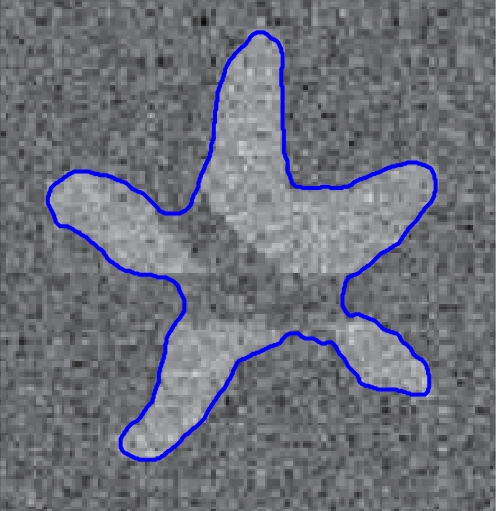}\\
\hline  
\end{tabular}
\caption{First column: Original image. Second column: results obtained from the ICTM for CV. Third column: results obtained from the ICTM for LIF. Fourth column: results obtained from the proposed TP-ICTM for CV. See Section~\ref{sec:comp2ICTM}.}
\label{fig:numerical_result}
\end{figure}

\begin{figure}
\centering
\begin{tabular}{|c|c|c|c|}
\hline
Initial guess & ICTM-CV & ICTM-LIF & TP-ICTM-CV\\
\hline
\includegraphics[width=0.16\linewidth]{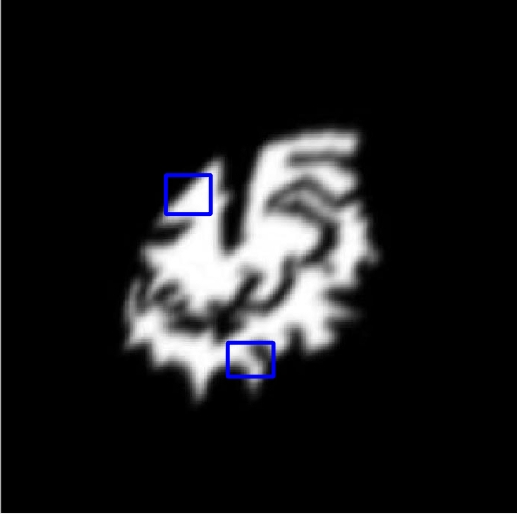}&
\includegraphics[width=0.16\linewidth]{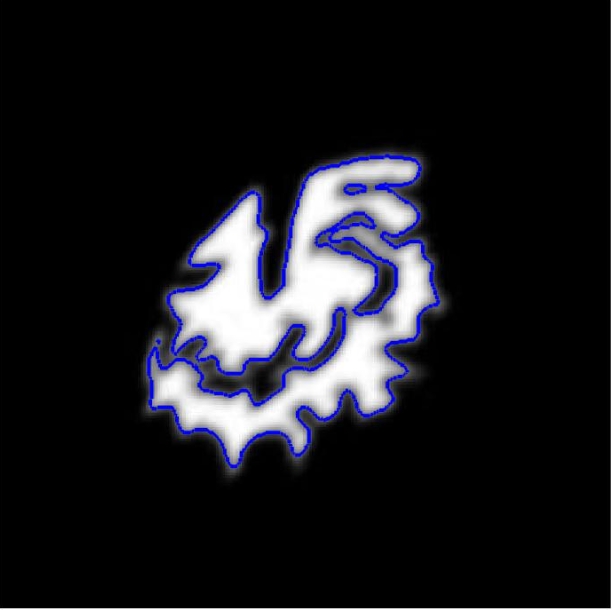}&
\includegraphics[width=0.16\linewidth]{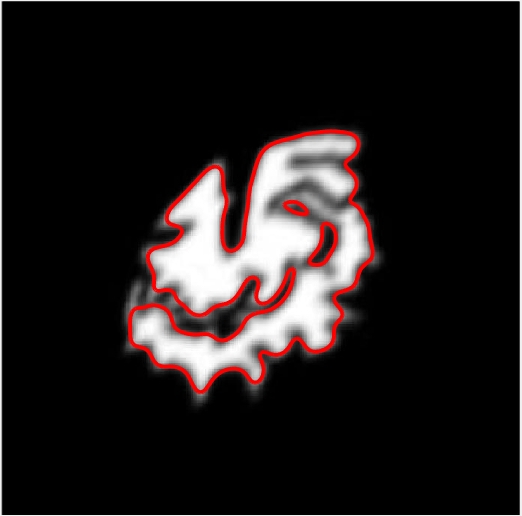}&
\includegraphics[width=0.16\linewidth]{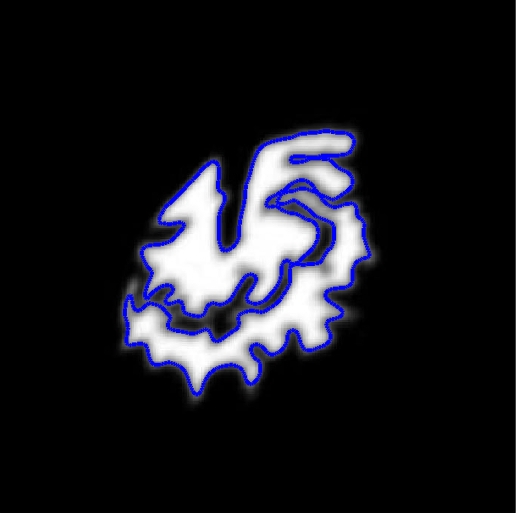}\\
\hline
Detail View &\includegraphics[width=0.16\linewidth]{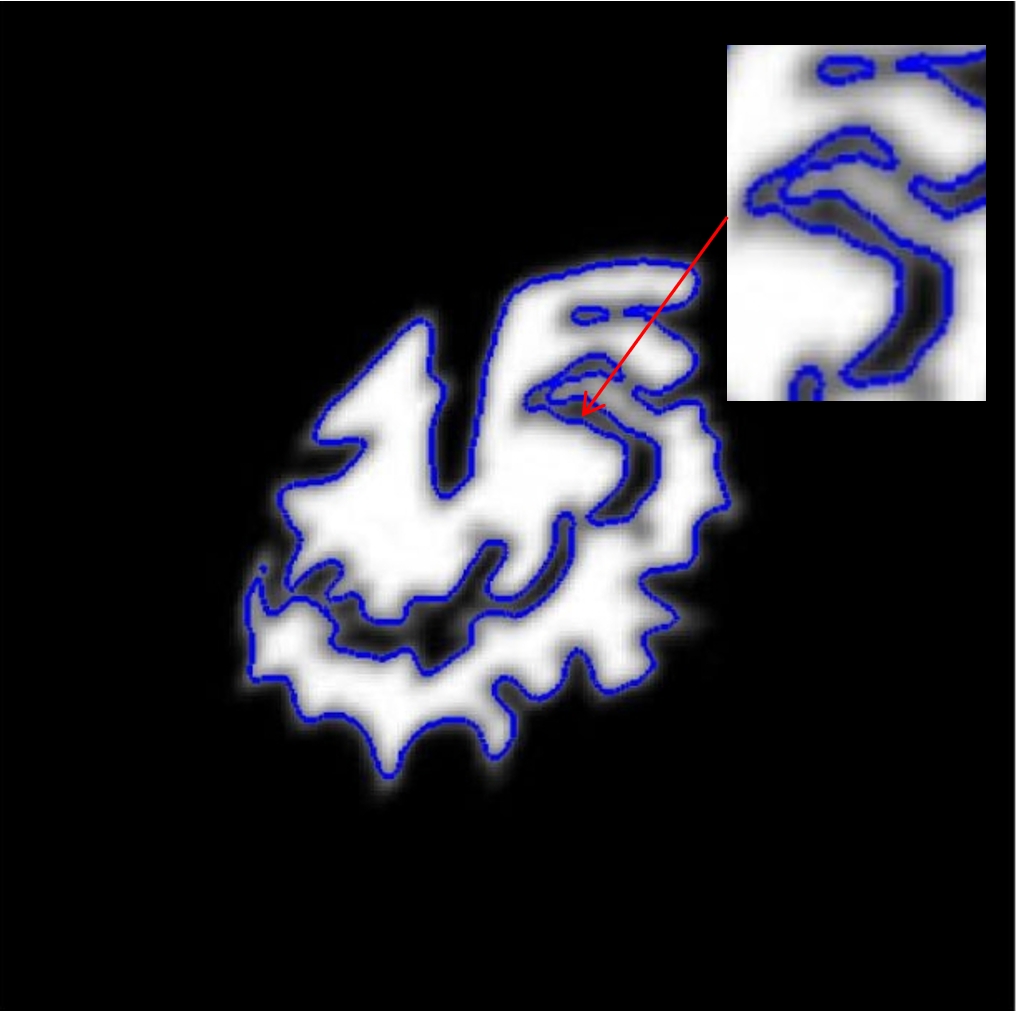}&
\includegraphics[width=0.16\linewidth]{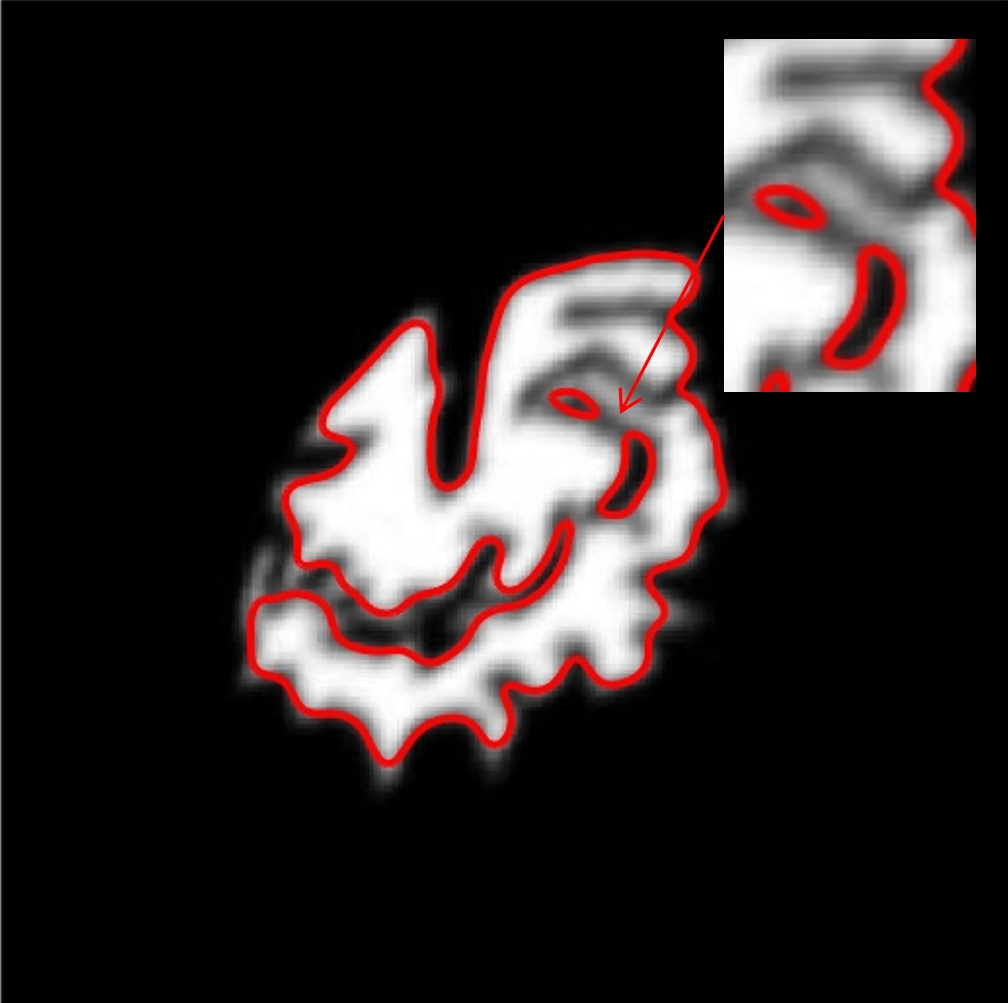}&
\includegraphics[width=0.16\linewidth]{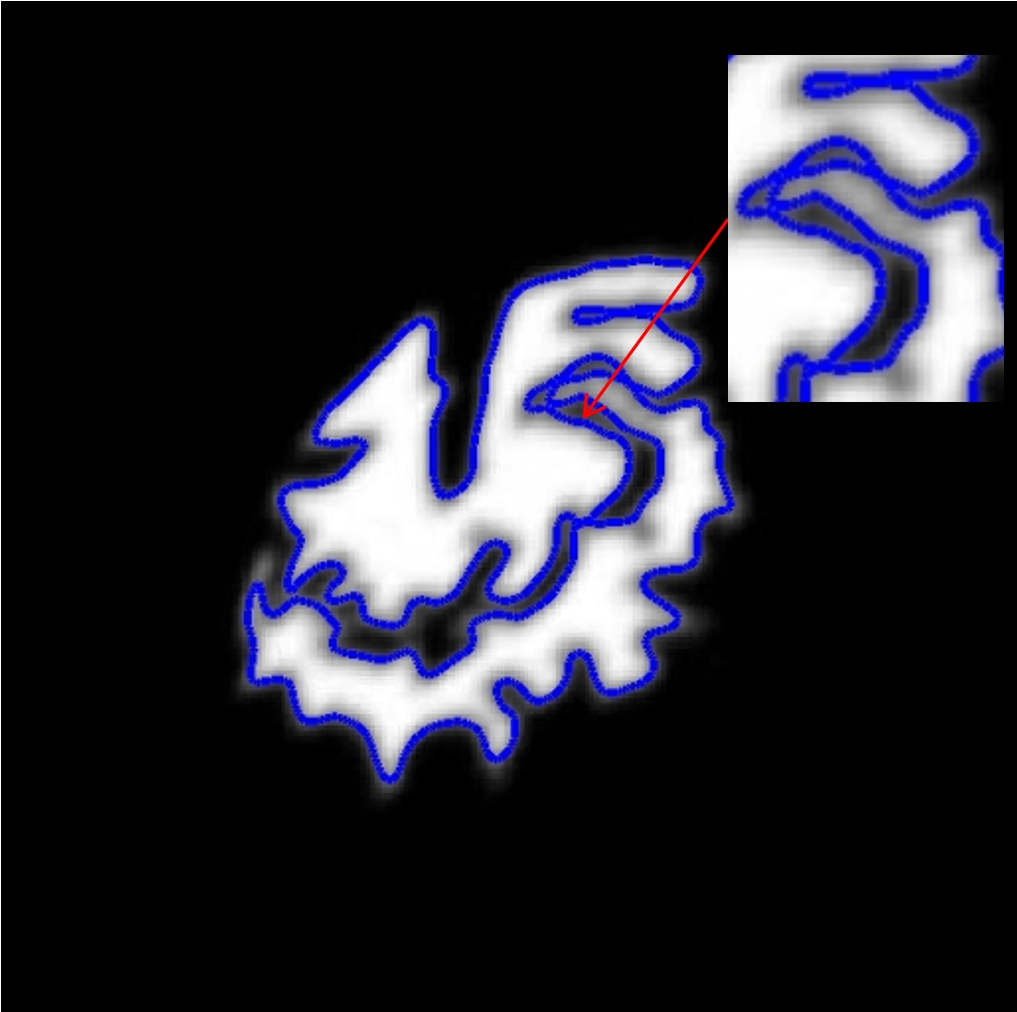}\\
\hline
\includegraphics[width=0.16\linewidth]{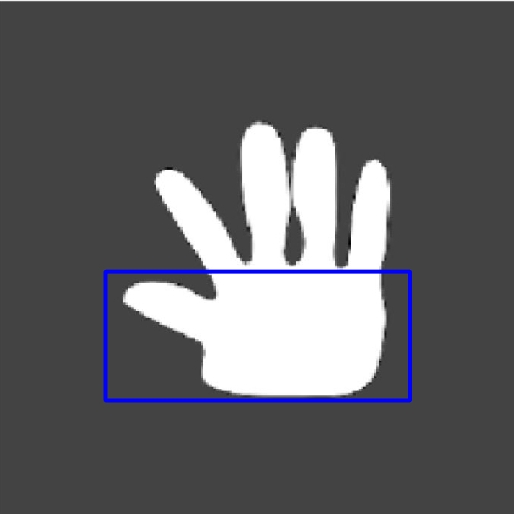}&  
\includegraphics[width=0.16\linewidth]{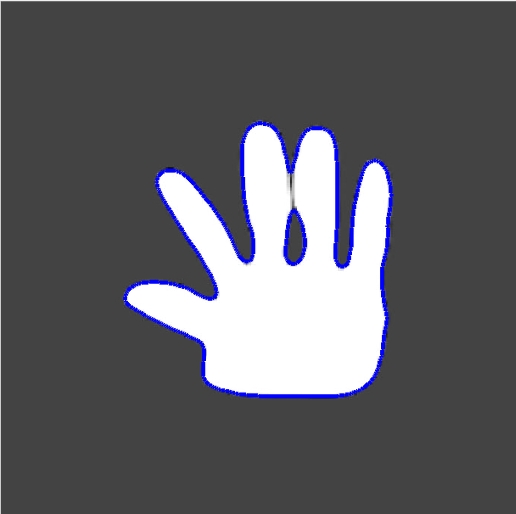}&
\includegraphics[width=0.16\linewidth]{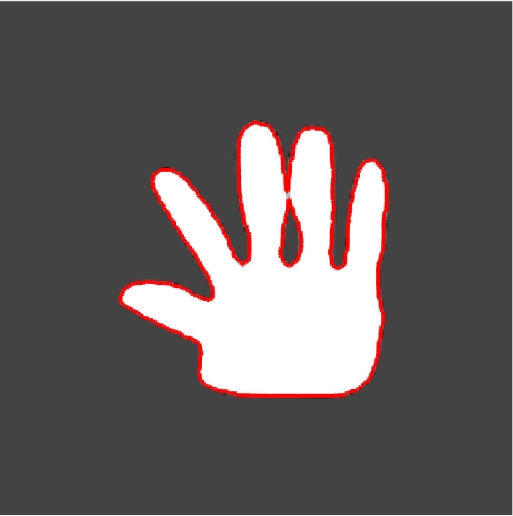}&
\includegraphics[width=0.16\linewidth]{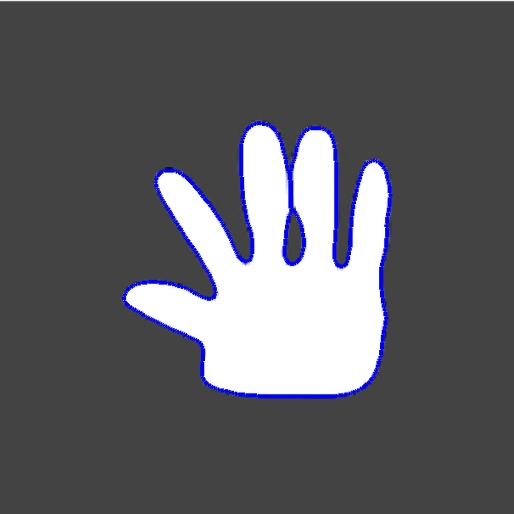}\\
\hline
Detail View & \includegraphics[width=0.16\linewidth]{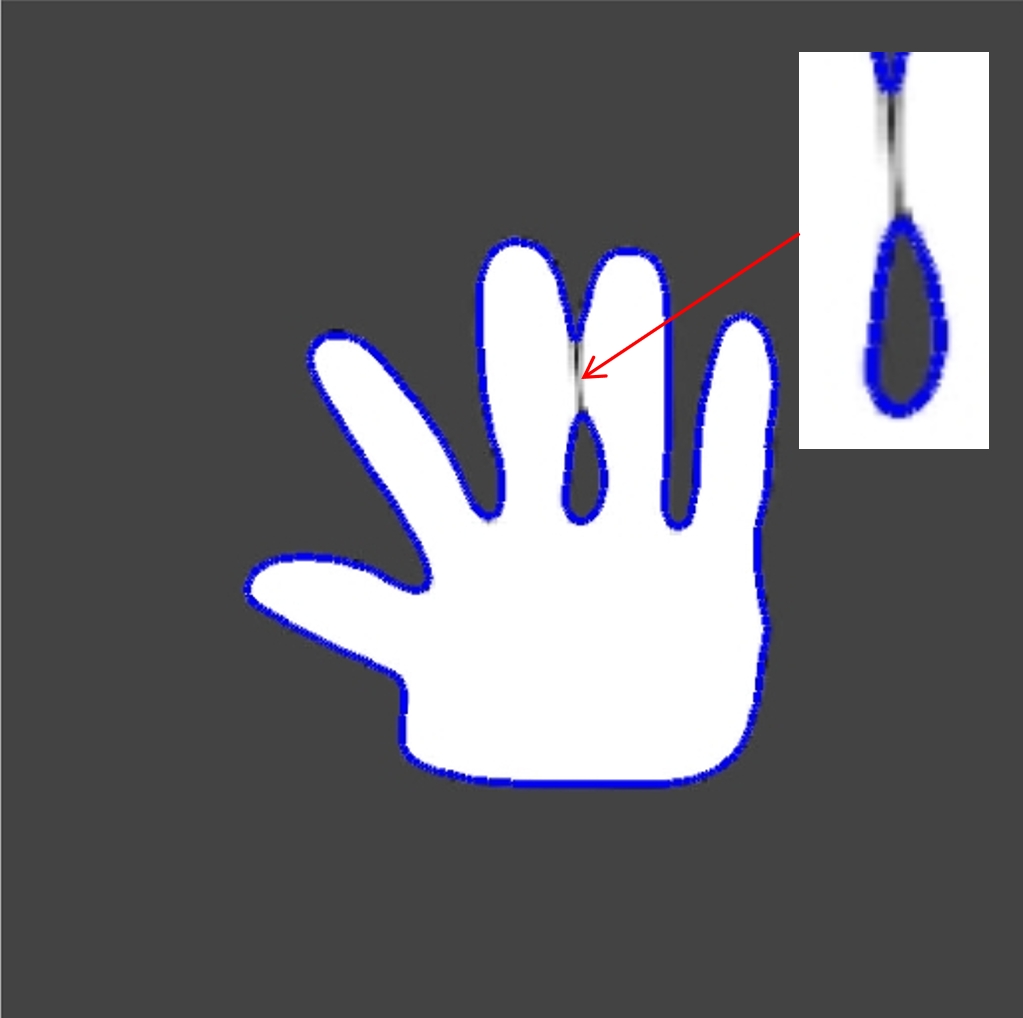}&
\includegraphics[width=0.16\linewidth]{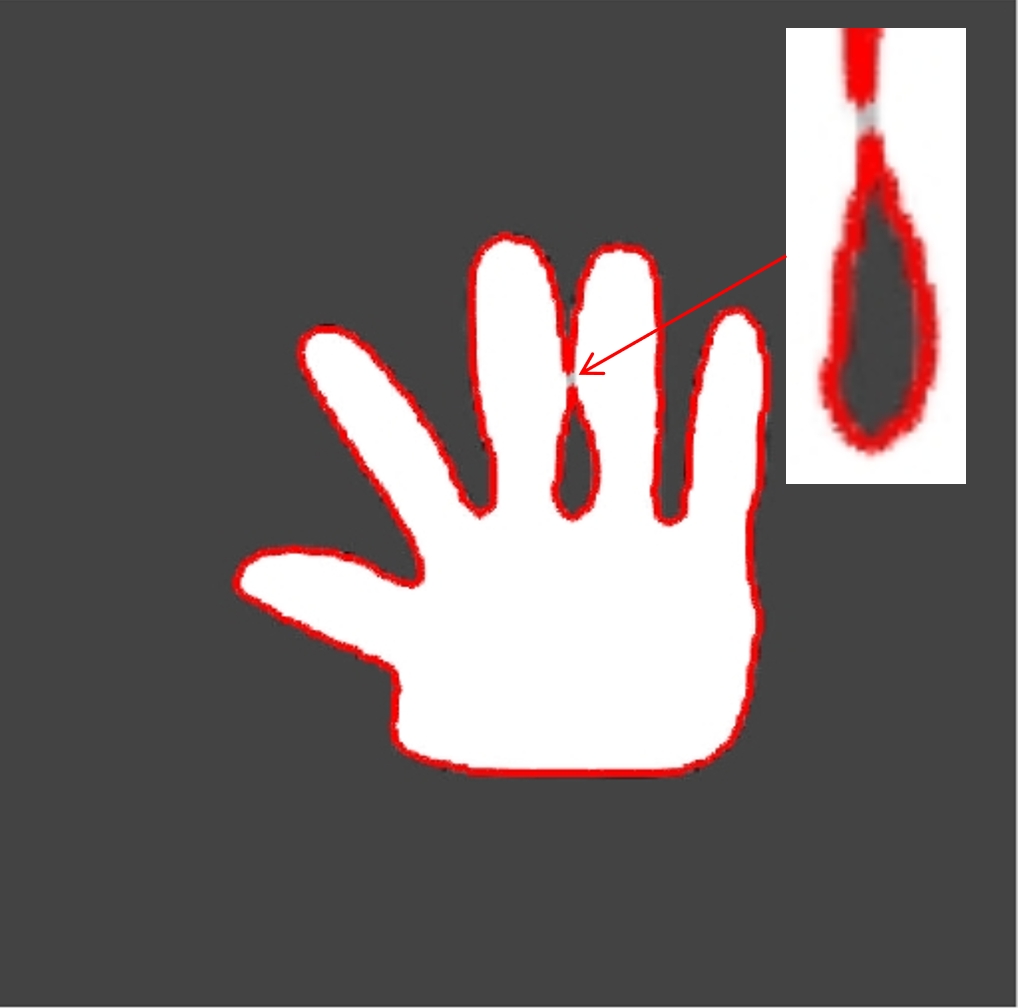}&
\includegraphics[width=0.16\linewidth]{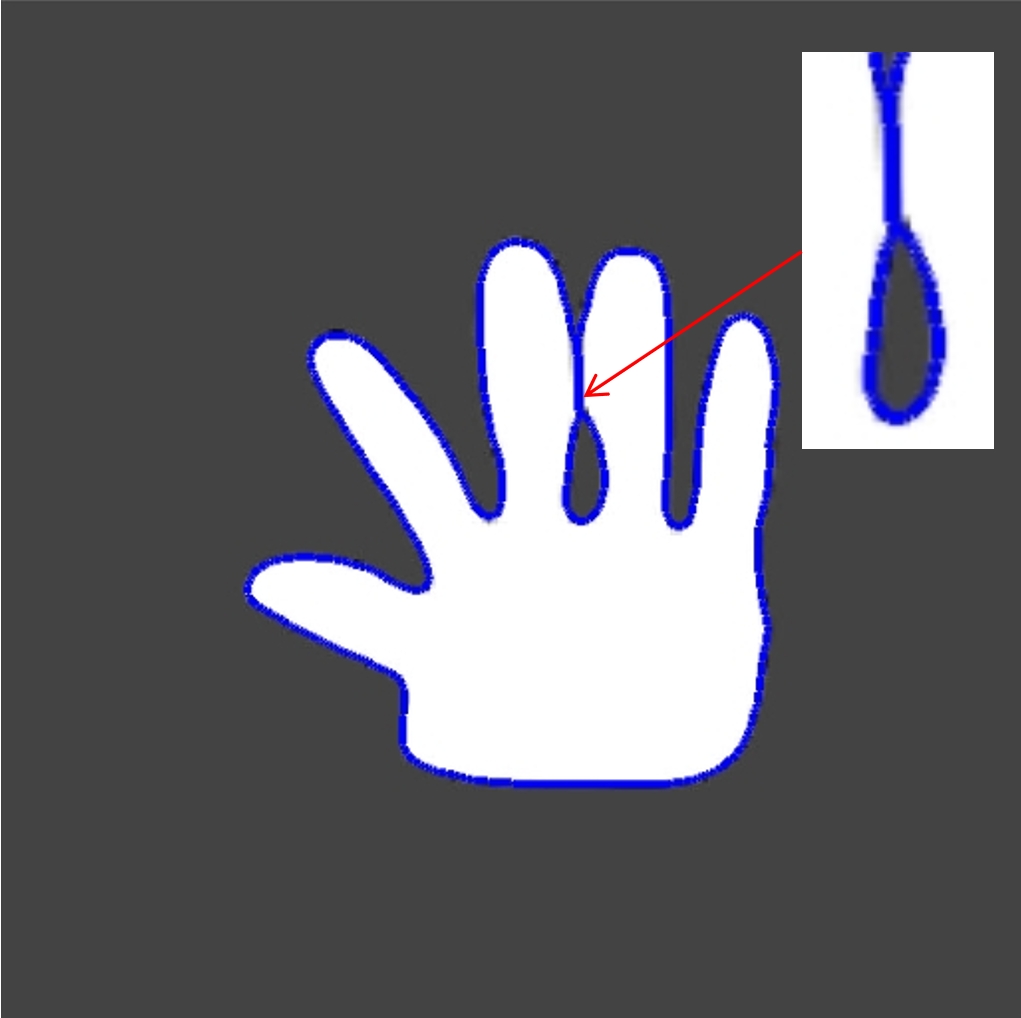}\\
\hline
\end{tabular}\caption{First column: original images with initial guesses. Second column: results obtained from the ICTM for CV. Third column: results obtained from the ICTM for LIF. Fourth column: results obtained from the proposed TP-ICTM for CV. Second and fourth row: detailed comparison of the results between ICTM for CV and TP-ICTM for CV in handling thin disconnected regions.
See Section~\ref{sec:comp2ICTM}.}
\label{fig:zoomoutcomparision}
\end{figure}

In this section, we demonstrate the differences between the proposed TP-ICTM for the CV model and the original ICTM \cite{wang2022iterative} for both the CV and LIF models. From the representative results shown in Figures~\ref{fig:numerical_result} and \ref{fig:zoomoutcomparision}, it can be observed that, in addition to preserving topological constraints, the TP-ICTM is significantly less sensitive to noise and patterns compared to the original ICTM. For instance, in the first three rows, TP-ICTM accurately produces a single-shaped result without any internal patterns; in the fourth row, it successfully generates a star-shaped result that preserves the topology, achieving an effect similar to image inpainting. Figure~\ref{fig:zoomoutcomparision}, TP-ICTM can obtain more accurate solutions by persevering the topology. In particular, the TP-ICTM for CV outperforms the ICTM for CV in scenarios involving thin disconnected parts by preserving topological integrity and ensuring accurate segmentation. In addition, for all examples shown in Figures~\ref{fig:numerical_result} and \ref{fig:zoomoutcomparision}, the parameters $\tau_2$ and $\lambda$ in TP-ICTM are the same as those in the ICTM for CV (i.e., $\tau$ and $\lambda$ in \cite{wang2022iterative}). This demonstrates the effectiveness of TP-ICTM and also suggests that TP-ICTM can be seamlessly integrated with other models.

\begin{figure}[ht!]
\centering
\includegraphics[height=0.11\textheight]{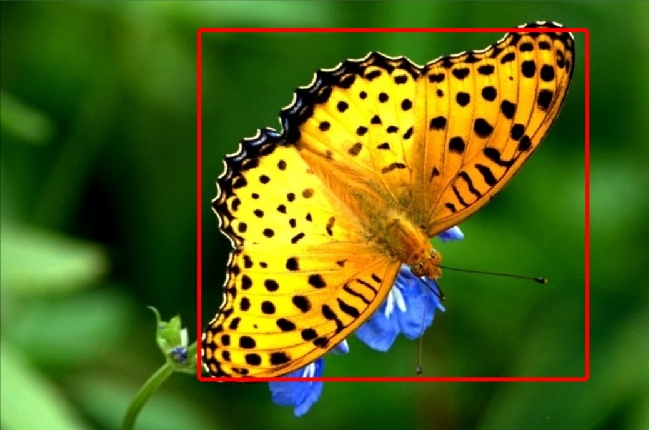}
\includegraphics[height=0.11\textheight]{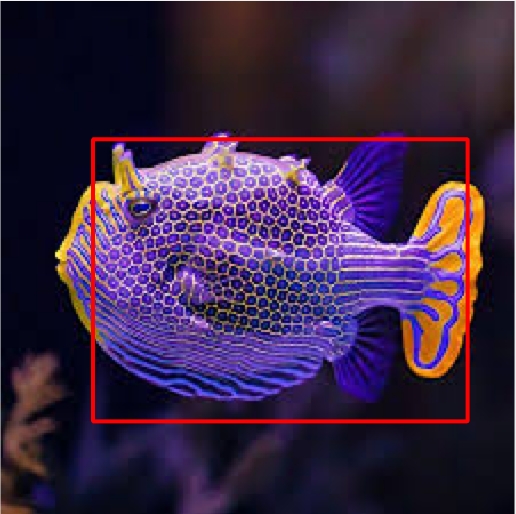}
\includegraphics[height=0.11\textheight]{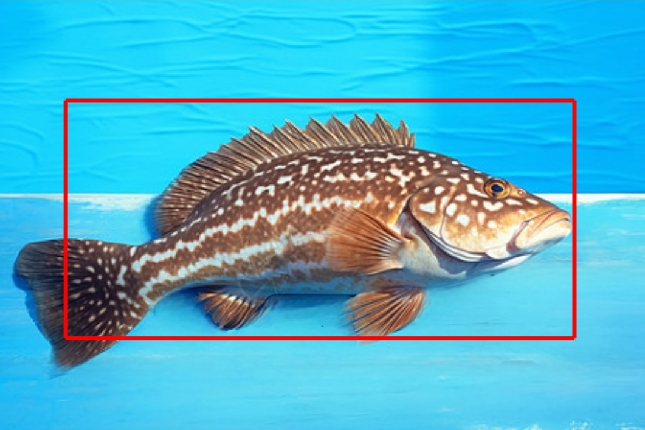}
\includegraphics[height=0.11\textheight]{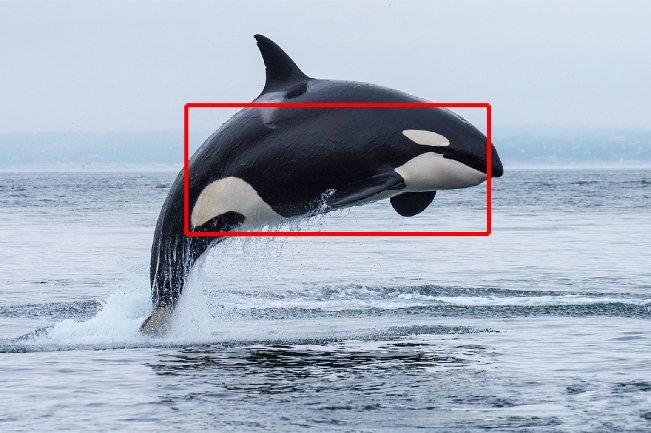}\\
\includegraphics[height=0.11\textheight]{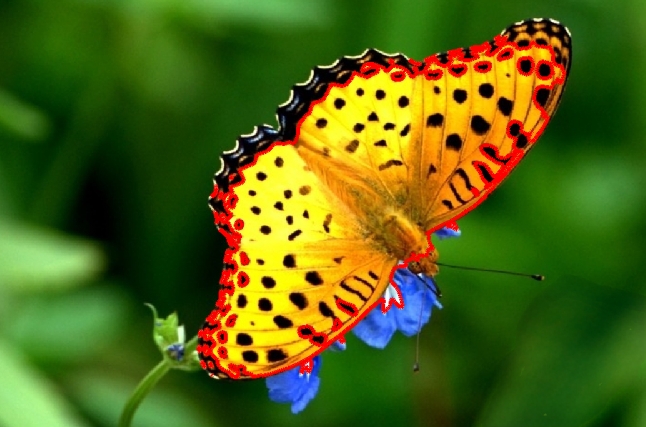}
\includegraphics[height=0.11\textheight]{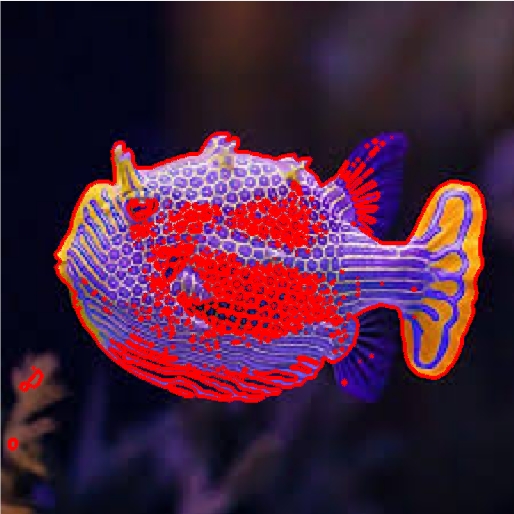}
\includegraphics[height=0.11\textheight]{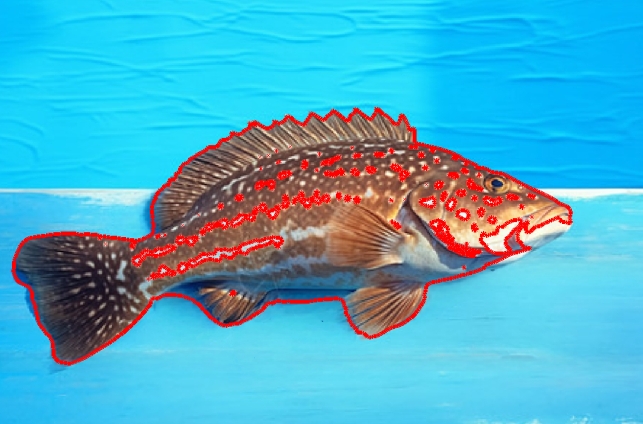}
\includegraphics[height=0.11\textheight]{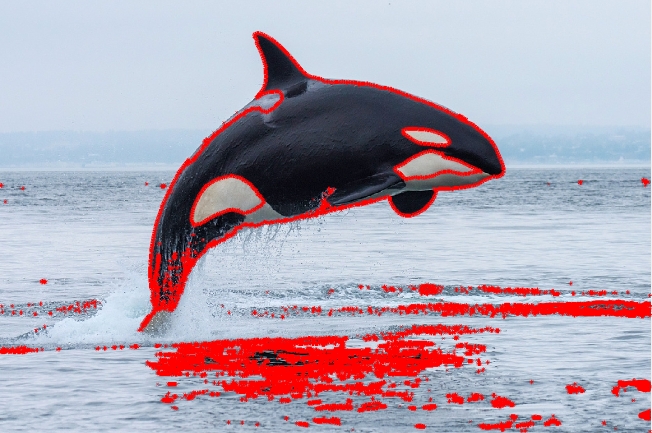}\\
\includegraphics[height=0.11\textheight]{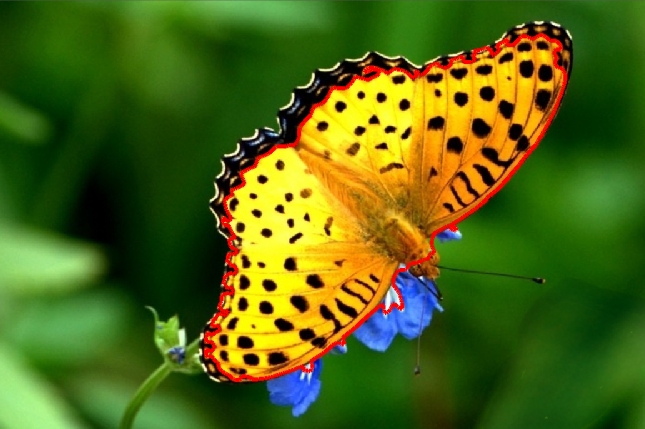}
\includegraphics[height=0.11\textheight]{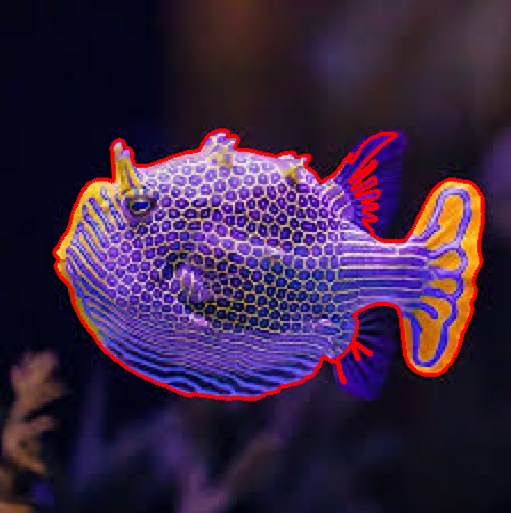}
\includegraphics[height=0.11\textheight]{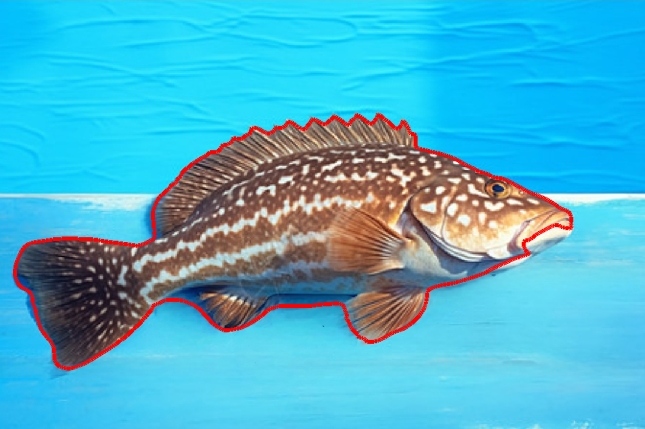}
\includegraphics[height=0.11\textheight]{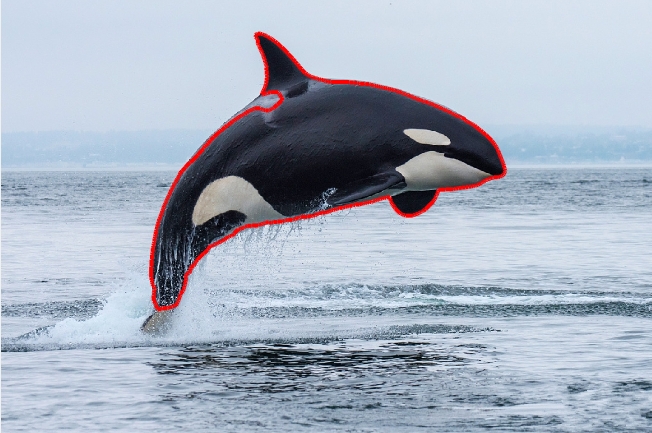}\\
\caption{First row: original image with initial guess. Second row: results obtained from ICTM-CV, number of iterations required is 89, 14, 45, and 23, respectively. Third row: results obtained from TP-ICTM-CV, number of iterations required is 72, 35, 63, and 89, respectively. See Section~\ref{sec:comp2ICTM}.}
\label{fig:color_result}
\end{figure}

In the second row of Figure~\ref{fig:color_result}, we present the results obtained using ICTM for CV for color images. The number of iterations required for each image is 89, 14, 45, and 23, respectively. The third row displays the results obtained using TP-ICTM-CV, with the number of iterations being 72, 35, 63, and 89, respectively. While TP-ICTM-CV may require slightly more iterations in some cases, it is particularly effective for images containing internal patterns. These patterns, which often include textures or repetitive structures, are accurately captured and segmented by TP-ICTM-CV, showcasing its robustness and adaptability. 

\begin{figure}[ht!]
\centering
\includegraphics[width=0.28\linewidth]{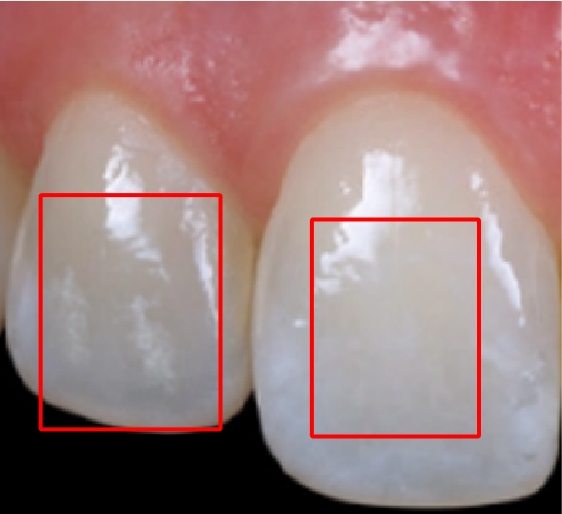}
\includegraphics[width=0.28\linewidth]{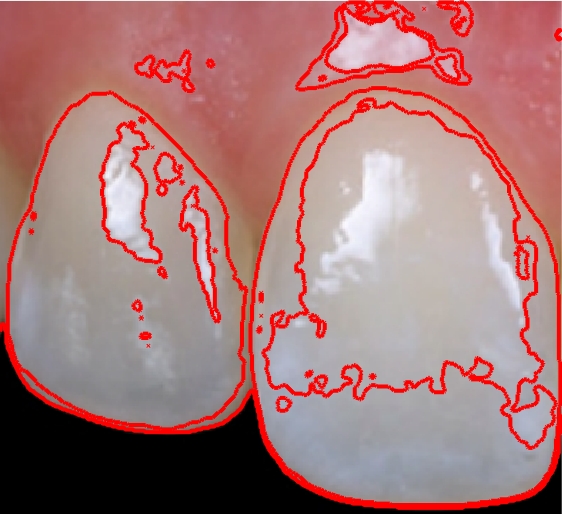}
\includegraphics[width=0.28\linewidth]{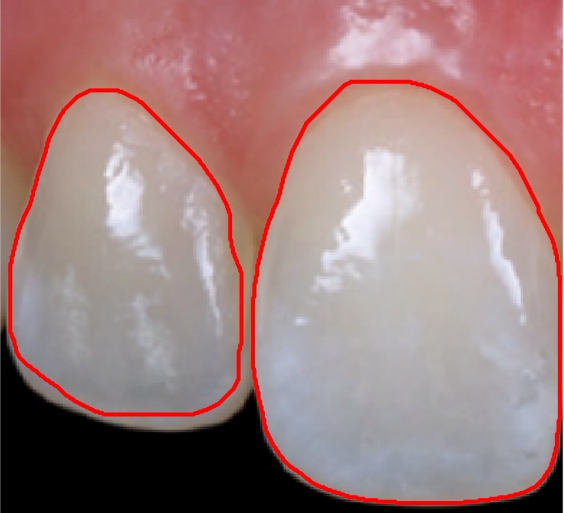}\\
\includegraphics[width=0.28\linewidth]{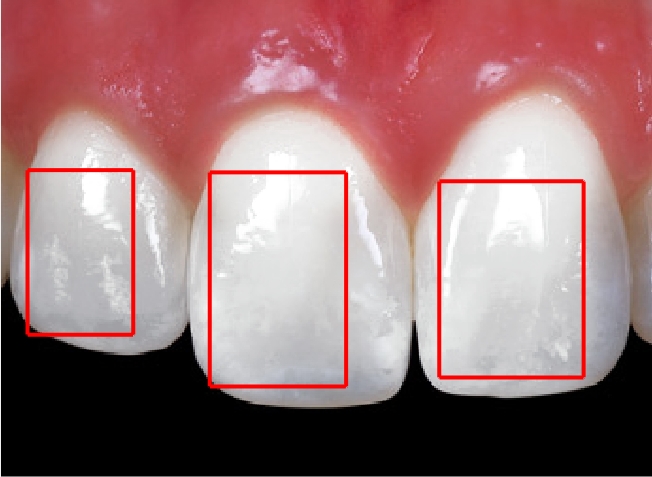}
\includegraphics[width=0.28\linewidth]{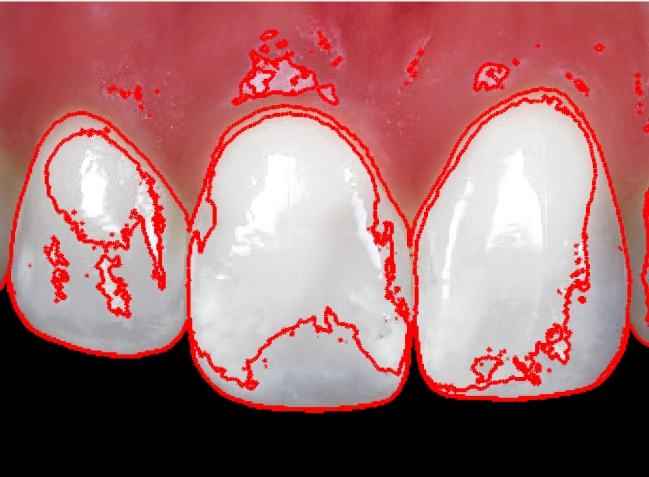}
\includegraphics[width=0.28\linewidth]{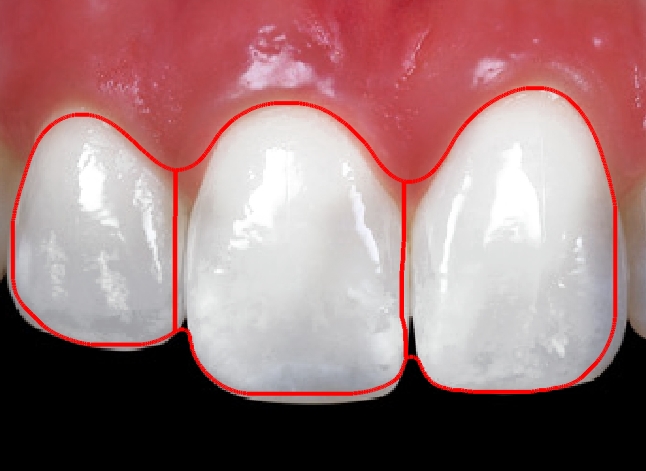}\\
\caption{First column: original image with initial guesses. Second column: results obtained from ICTM-CV. Third column: results obtained from TP-ICTM-CV. See Section~\ref{sec:comp2ICTM}.}
\label{fig:teeth}
\end{figure}

In Figure~\ref{fig:teeth}, we present the results obtained using TP-ICTM-CV for the human teeth image analyzed in \cite{kim2024individual}. Segmenting individual teeth in images captured outside the oral cavity has challenges including weak edges, intensity inhomogeneities, and strong light reflections. The proposed TP-ICTM-CV method effectively addresses these issues, generating smooth and accurate segmentation lines that closely align with the actual teeth, outperforming the ICTM-CV approach. Notably, TP-ICTM demonstrates robustness against light reflections on teeth, further highlighting its reliability.

\subsection{Application to LIF model}\label{sec:LIF}
In this section, we extend the application of the proposed algorithms into the LIF model \cite{li2008minimization} where $F_i=\lambda_i\int_{\Omega}G_{\delta}(x-y)|C_i (x)-f(y)|^2dx$ and $\Theta_i =C_i (x)$ for $i=1,2$.
When $u^k$ is fixed,\begin{align*}
\lambda_1\int_{\Omega} G_{\tau_1} *u^k\int_{\Omega}G_{\delta}(x-y)|C_1 (x)-f(y)|^2 \ dx \ dy \\+\lambda_2 \int_{\Omega} (1-G_{\tau_1} *u^k)\int_{\Omega}G_{\delta}(x-y)|C_2 (x)-f(y)|^2 \ dx\ dy
\end{align*}
is strictly convex with respect to $C_1(x)$ and $C_2(x)$. Then, according to variation with respect to $C_1(x)$ and $C_2(x)$, we have \begin{align*}
    \int \int_{\Omega} (G_{\tau_1}*u^k)(y) G_{\delta} (x-y)[C_1(x)-f(y)]dydx =0
  \\\int \int_{\Omega}(G_{\tau_1}*(1- u^k))(y)G_{\delta} (x-y)[C_2(x)-f(y)]dydx =0
\end{align*}
to obtain\begin{align*}
C^k_1=\frac{G_{\delta} *(G_{\tau_1}*u^k f)}{G_{\delta} *G_{\tau_1}*u^k},
C^k_2=\frac{G_{\delta} *G_{\tau_1}*((1-u^k) f)}{G_{\delta}*G_{\tau_1}*(1-u^k)}.
\end{align*}

The algorithm is summarized below in Algorithm~\ref{alg:topology preserving LIF}.

\begin{rem}
There may exist some $x$ such that $(G_{\delta} *G_{\tau_1}*u^k)(x)$ or $(G_{\delta} *G_{\tau_1}*(1-u^k))(x)$ is close to $0$. In the practical implementation, with a small parameter $\varepsilon>0$, we use 
\begin{align*}
C^k_1=\frac{G_{\delta} *(G_{\tau_1}*u^k f)}{G_{\delta} *G_{\tau_1}*u^k+\varepsilon},
C^k_2=\frac{G_{\delta} *G_{\tau_1}*((1-u^k) f)}{G_{\delta}*G_{\tau_1}*(1-u^k)+\varepsilon}.
\end{align*}
\end{rem}

\begin{algorithm}[!htbp]
\caption{TP-ICTM for LIF}
\label{alg:topology preserving LIF}
\begin{algorithmic}
\REQUIRE Given image $f$, $\tau_1, \tau_2, \delta > 0$; tolerance parameter $tol > 0$;  $u^0=\chi_{\Omega^0}$;.
\ENSURE Segmentation of the image $u$;
\WHILE{$||u^{k+1}-u^k||_2>tol$}
    \STATE 1. Compute \begin{equation*}
    C^k_1=\frac{G_{\delta} *((G_{\tau_1}*u^k) f)}{G_{\delta} *G_{\tau_1}*u^k}, \ \ \ \ \ C^k_2=\frac{G_{\delta} *((G_{\tau_1}*(1-u^k)) f)}{G_{\delta}*G_{\tau_1}*(1-u^k)}.\end{equation*}
        
    \STATE 2. Compute $\phi^k(x) = G_{\tau_1} *(F_1^k-F_2^k) + \mu \sqrt{\frac{\pi}{\tau}} G_{\tau_2} *(1-2u^k).$\;
    
    \STATE 3. Thresholding: \begin{equation}\nonumber
        u^{*}(x) = \begin{cases}1 \ \  \textrm{if}  \ \phi^k(x)\leq 0 ,   \\
0 \ \  \textrm{otherwise}.  \end{cases}
    \end{equation}
    \STATE 4. Topology preserving: $u^{k+1} = \textnormal{KeepConnectivity}(u^k, u^{*}, \phi^k).$ \;
     \STATE $k=k+1$.    
\ENDWHILE
\end{algorithmic}
\end{algorithm}

\begin{figure*}[ht!]
\centering
\includegraphics[width=0.16\linewidth]{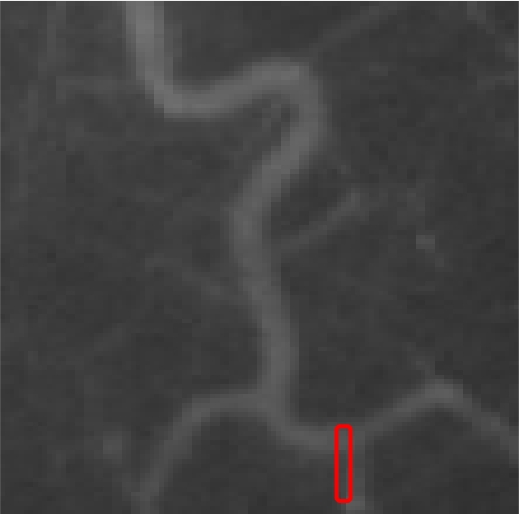}
\includegraphics[width=0.16\linewidth]{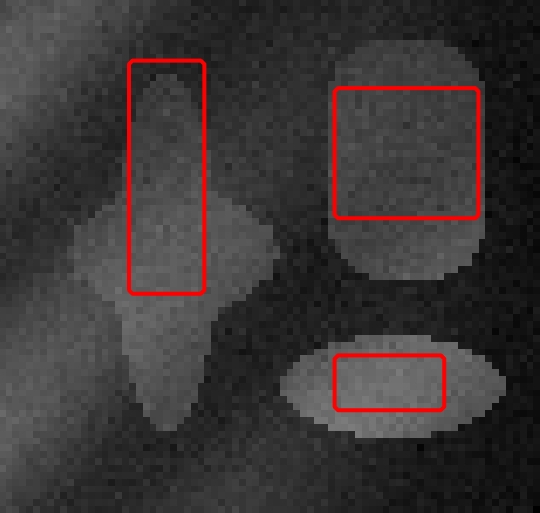}
\includegraphics[width=0.16\linewidth]{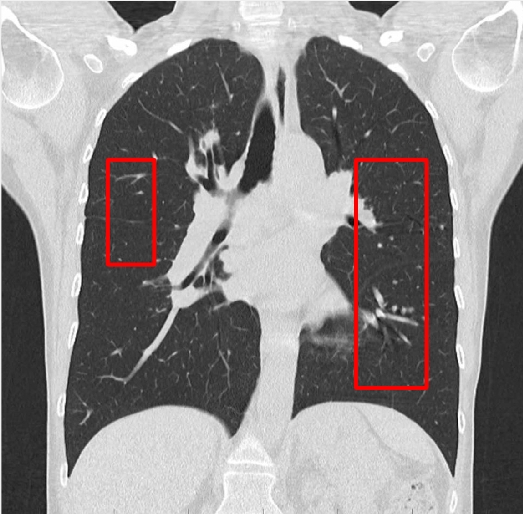}
\includegraphics[width=0.16\linewidth]{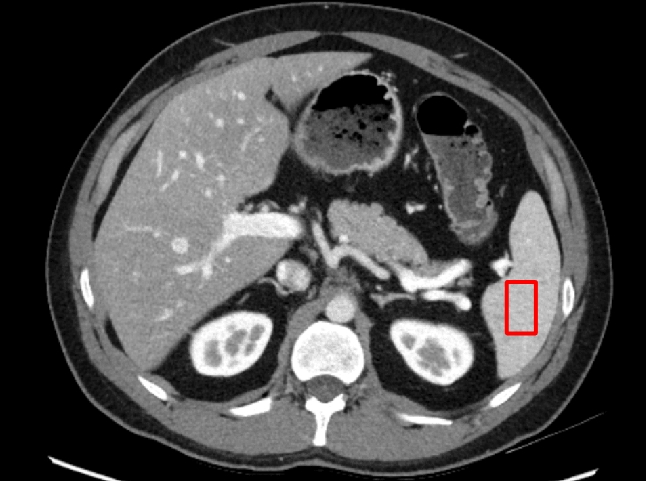} \includegraphics[width=0.16\linewidth]{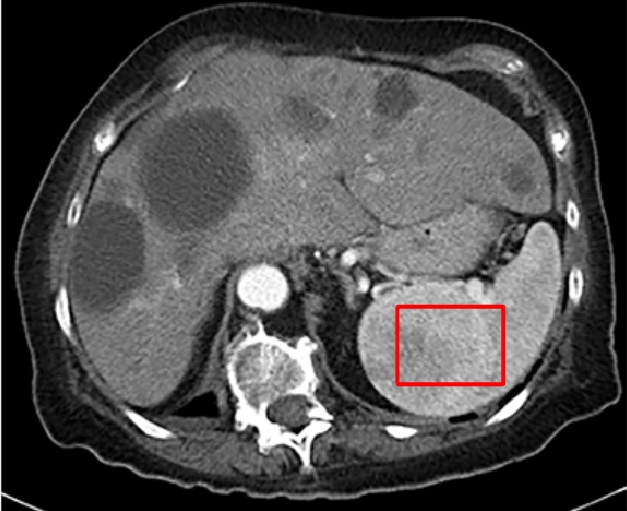}\\
\includegraphics[width=0.16\linewidth]{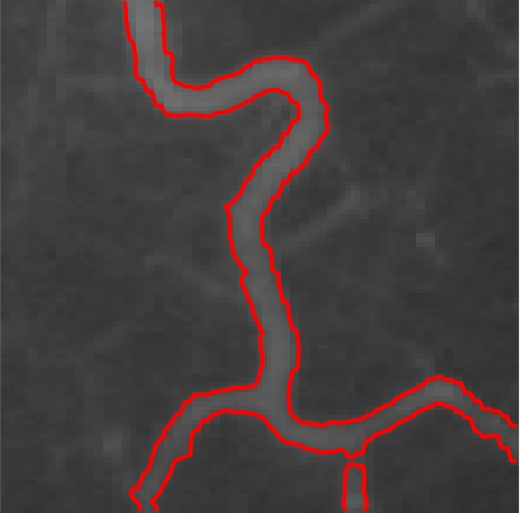}
\includegraphics[width=0.16\linewidth]{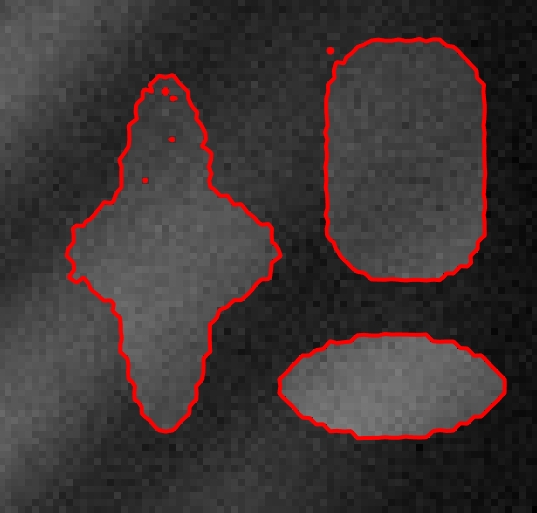}
\includegraphics[width=0.16\linewidth]{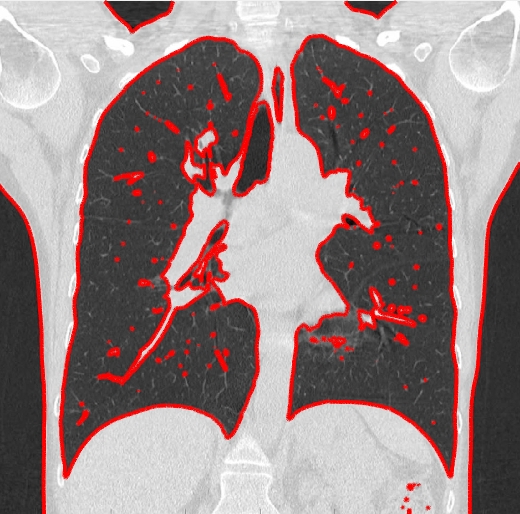}
\includegraphics[width=0.16\linewidth]{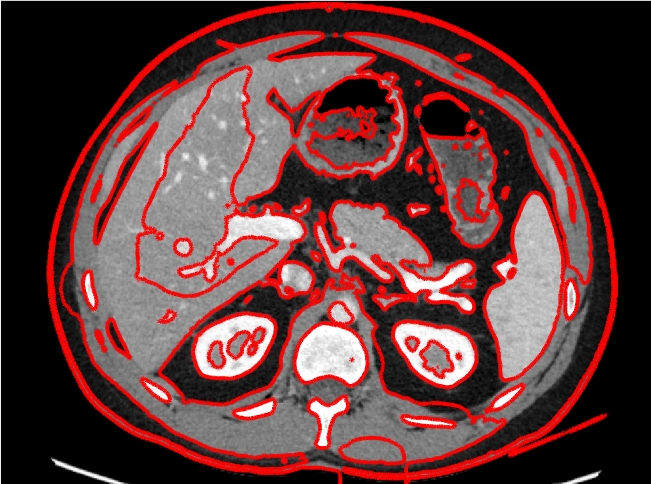}
\includegraphics[width=0.16\linewidth]{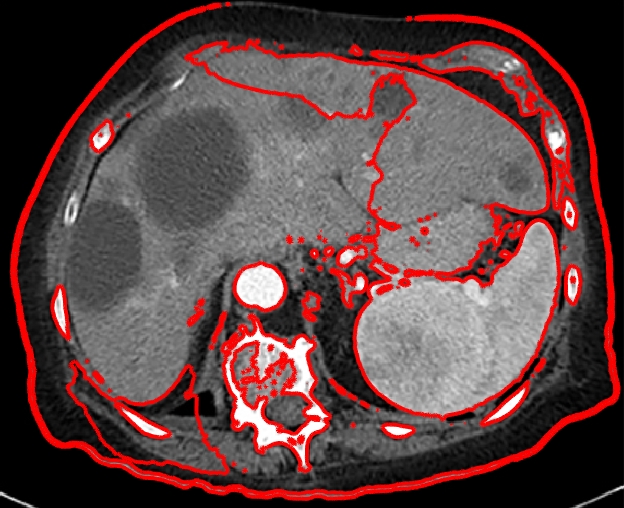}\\
\includegraphics[width=0.16\linewidth]{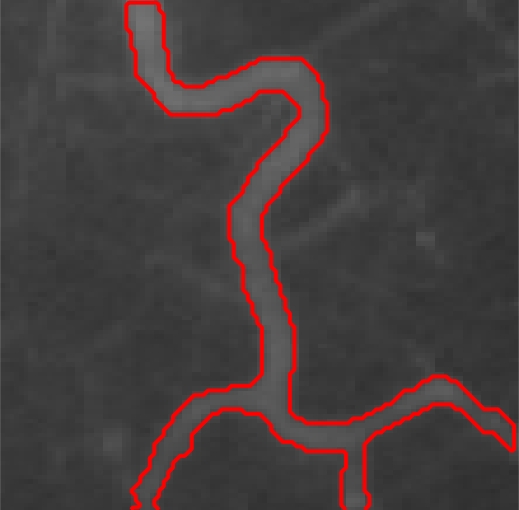}
\includegraphics[width=0.16\linewidth]{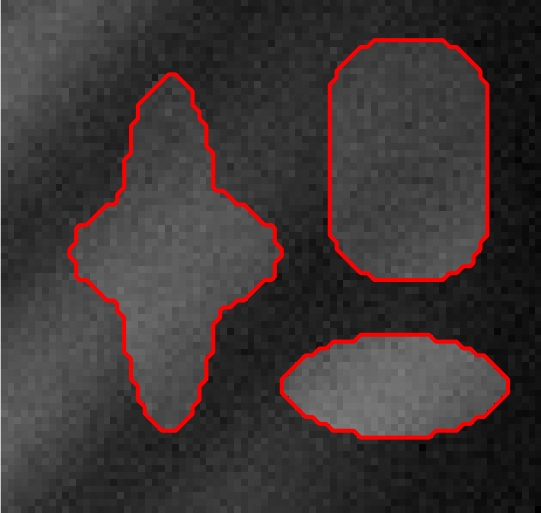}
\includegraphics[width=0.16\linewidth]{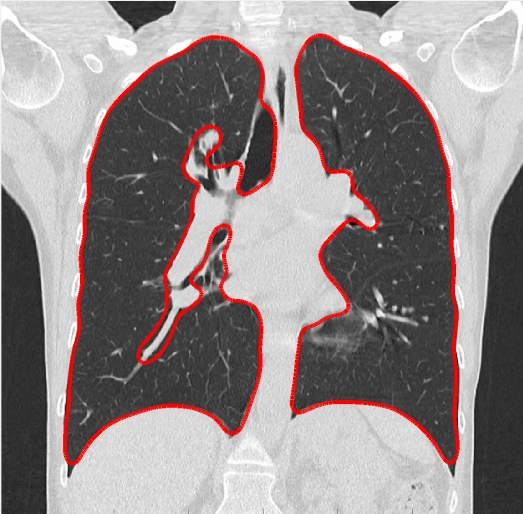}
\includegraphics[width=0.16\linewidth]{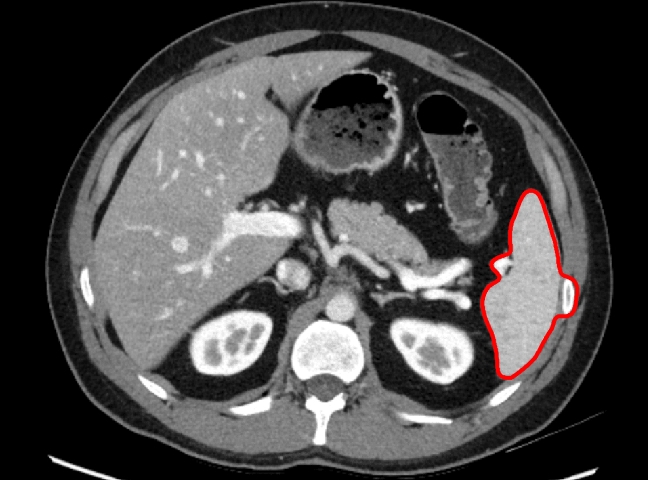}
\includegraphics[width=0.16\linewidth]{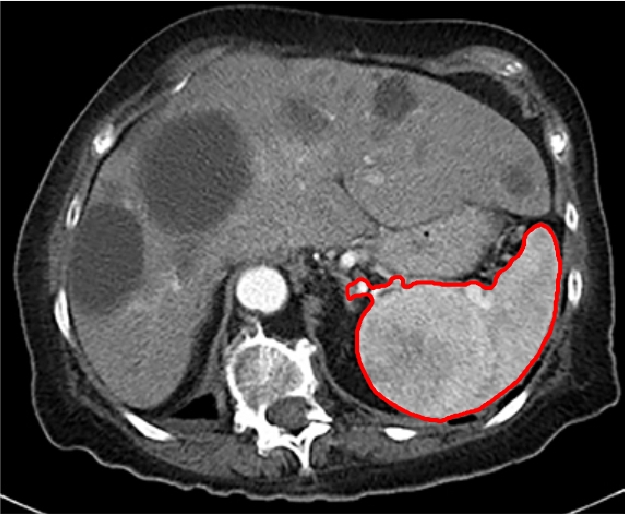}\\
\caption{First row: Initial guesses, second row: results obtained from ICTM for LIF, third row: results obtained from TP-ICTM for LIF. See Section~\ref{sec:LIF}.}
\label{fig:LIF_model_extension}
\end{figure*}

As illustrated in Figure~\ref{fig:LIF_model_extension}, the application of the algorithm to the LIF model demonstrates its effectiveness in addressing challenges such as discontinuities and noise that deviate from the ground truth topological properties (shown in the first to third columns of Figure~\ref{fig:LIF_model_extension}). Additionally, the algorithm successfully segments the specified regions of interest. Notably, the last two columns of Figure~\ref{fig:LIF_model_extension} highlight its performance on real medical spleen-CT images, where the algorithm leverages topological properties to accurately isolate and segment the spleen.

Regarding parameters, the TP-ICTM and ICTM methods utilize identical settings, consistent with those used in the comparative analysis with the CV model, except for the inclusion of the additional parameter $\tau_1$. The consistent performance across these methods underscores the robustness and effectiveness of the proposed approach.

In terms of efficiency, the ICTM required $49$, $9$, $19$, $157$, and $277$ iterations, while the TP-ICTM required $58$, $16$, $171$, $79$, and $119$ iterations, respectively. In most cases, the TP-ICTM demonstrates comparable efficiency to the ICTM.

\section{Conclusion and discussions}\label{sec:con}
In this paper, we proposed a novel approach for minimizing variational models for image segmentation with topology preservation constraints, leveraging the framework of ICTM. The proposed TP-ICTM extends ICTM by incorporating a topology-checking step, ensuring the preservation of desired topological structures. This enhancement retains the core advantages of ICTM—simplicity, computational efficiency, and numerical stability—while significantly improving robustness, particularly in handling noisy images or those with complex internal patterns.
Extensive numerical experiments demonstrate that TP-ICTM outperforms the standard ICTM in scenarios where topological accuracy is critical. Moreover, the flexibility of TP-ICTM enables seamless integration with other segmentation models or methods that require topology constraints. This adaptability positions TP-ICTM as a promising tool for applications such as human-machine interaction in image segmentation, where maintaining specific topological priorities is crucial.
Future work could focus on extending TP-ICTM to other domains, such as topology optimization \cite{chen2024prediction}, or on developing more advanced topology-preserving mechanisms to dynamically adapt or select appropriate topologies during optimization. 

\bibliographystyle{plain} 
\bibliography{ref.bib} 

\end{document}